\documentclass[10pt,twocolumn,letterpaper]{article}

\usepackage[pagenumbers]{cvpr} 

\usepackage[T1]{fontenc}
\usepackage{cuted} 

\usepackage{currfile} 

\usepackage{caption} 

\usepackage{xspace}

\usepackage{amsthm}
\usepackage{multirow}
\usepackage{colortbl}
\usepackage{tabularx}
\usepackage{algorithm}
\usepackage{algpseudocode}

\definecolor{vbbestbg}{RGB}{168,218,181}
\definecolor{vbsecondbg}{RGB}{226,242,230}
\definecolor{vbthirdbg}{RGB}{254,243,199}
\newcommand{\vbbest}[1]{\cellcolor{vbbestbg}\textbf{#1}}
\newcommand{\vbsecond}[1]{\cellcolor{vbsecondbg}\underline{#1}}
\newcommand{\vbthird}[1]{\cellcolor{vbthirdbg}#1}
\usepackage{tikz}
\usetikzlibrary{arrows.meta}
\newlength{\timearrowlen}

\makeatletter
\def\theHALG@line{\arabic{algorithm}.\arabic{ALG@line}}
\makeatother

\definecolor{rankone}{HTML}{0A327A}
\definecolor{ranktwo}{HTML}{2E58A3}
\definecolor{rankthree}{HTML}{537FCD}
\definecolor{rankother}{HTML}{7EA5EA}
\definecolor{rankref}{HTML}{808080}

\newcommand{\unranked}{\color{rankref}}

\newtheorem{proposition}{Proposition}

\newcommand{\piflow}{$\pi$-Flow\xspace}

\newcommand{\methodone}{anti-drifting guidance\xspace}
\newcommand{\Methodone}{Anti-Drifting Guidance\xspace}
\newcommand{\methodtwo}{motion-aware guidance\xspace}
\newcommand{\Methodtwo}{Motion-Aware Guidance\xspace}
\newcommand{\sg}{\operatorname{sg}}
\algrenewcommand\algorithmiccomment[1]{\hfill{\small\color{gray}$\triangleright$ #1}}

\definecolor{cvprblue}{rgb}{0.21,0.49,0.74}
\usepackage[pagebackref,breaklinks,colorlinks,allcolors=cvprblue]{hyperref}

\title{Analyzing and Improving FlowMap Distillation of Autoregressive Video Models}

\author{Yong-Hyun Park\textsuperscript{1,$*$}\quad
Guande He\textsuperscript{2}\quad
Joon Hyeok Kim\textsuperscript{1}\quad
Hansheng Chen\textsuperscript{3}\\[1pt]
Kaifan Yu\textsuperscript{1}\quad
Chaoqi Luo\textsuperscript{1}\quad
Xun Huang\textsuperscript{2}\quad
Jiatao Gu\textsuperscript{1}\\[4pt]
\textsuperscript{1}University of Pennsylvania\quad
\textsuperscript{2}Roblox\quad
\textsuperscript{3}Stanford University\\[-8pt]
}

\begin{document}
\maketitle
\begingroup
\renewcommand\thefootnote{$*$}
\footnotetext{This work was partially done during an internship at Roblox. Correspondence: \href{mailto:park19@engineering.upenn.edu}{\texttt{park19@engineering.upenn.edu}}.}
\addtocounter{footnote}{-1}
\endgroup

\begin{abstract}
FlowMap distillation offers a natural route to few-step generative modeling by directly learning a teacher's denoising trajectories. Yet for autoregressive (AR) video generation, existing few-step models still rely on distribution matching distillation (DMD) to remain competitive, sacrificing sample diversity and requiring a large bidirectional teacher. In this work, we analyze what limits standalone AR FlowMap distillation and introduce two complementary guidance mechanisms to close this gap: (1) \textbf{\methodone}, which helps the AR teacher correct drifted student rollouts by training an auxiliary model to mimic student drift and using it as a negative guidance branch; and (2) \textbf{\methodtwo}, which makes motion easier for a FlowMap student to learn by locating motion patches and selectively applying skip-layer guidance. Empirically, our approach applies to \piflow, MeanFlow, and consistency distillation with 1.3B AR teachers, improving generation quality and reducing drift over 30\,s rollouts. Notably, our 4-step \piflow students match DMD students distilled from 14B teachers on VBench while preserving the sample diversity of their AR teachers, demonstrating that pure FlowMap distillation is effective for few-step autoregressive video generation.\\[4pt]
\noindent\textbf{Project page:} \url{https://enkeejunior1.github.io/ar-video-flowmap/}
\end{abstract}

\section{Introduction}
\label{sec:introduction}

Standard video diffusion models denoise all frames jointly with bidirectional attention \citep{ho2022video,blattmann2023stable,kong2024hunyuanvideo,hacohen2024ltx,wan2025wan}. While effective for offline synthesis, generating an entire video at once prevents streaming and real-time interaction. Autoregressive (AR) video diffusion models instead generate a video chunk by chunk, conditioning each chunk on the preceding ones \citep{chen2024diffusion,yin2025slow,teng2025magi}. This causal structure lets the model stream video and respond to user input, supporting interactive generation and world simulation \citep{li2025hunyuan,shin2026motionstream,yang2025longlive}. However, standard diffusion sampling takes many denoising steps per chunk, which is slow for real-time generation. Practical deployment therefore relies on distilling AR video models into students that generate each chunk in only a few steps \citep{yin2025slow,huang2026self,lin2026autoregressive}.

\begin{figure*}[t]
\centering
\includegraphics[width=\linewidth]{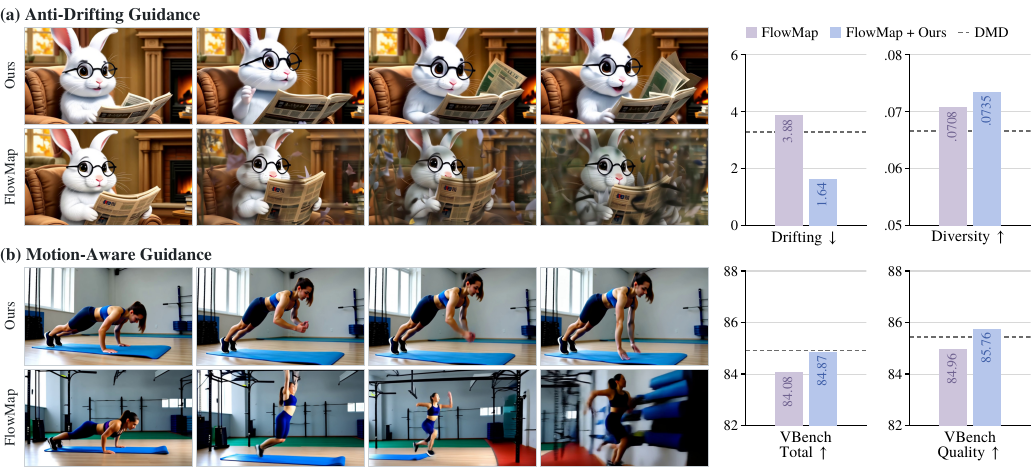}\\[-3pt]
\caption{\textbf{Comparison of AR FlowMap distillation with and without our guidance.} (a)~\Methodone reduces student drift over 30\,s video generation while preserving sample diversity. (b)~\Methodtwo makes motion easier for the student to learn, improving generation quality.}
\label{fig:\currfilebase}%
\vspace{-4pt}
\end{figure*}

Current methods perform this distillation in two stages \citep{huang2026self,zhu2026causal}. Stage~1 initializes the student by conditioning each chunk on clean context frames (\emph{teacher-forcing}), but training only on clean histories causes errors to accumulate across generated chunks at inference. To address this exposure bias, Stage~2 rolls out the student autoregressively during training (\emph{self-forcing}) and supervises the generated sequences with distribution matching distillation (DMD) \citep{yin2024one,yin2024improved,huang2026self}, which matches the student's video distribution to a pretrained teacher. While self-forcing with DMD (SF-DMD) mitigates rollout drift \citep{huang2026self,zhuang2026self,su2026sgf+} and has been adopted by most recent few-step AR video models \citep{zhu2026causal,zhao2026causal,zheng2026causal,yang2025longlive,liu2026rolling}, DMD reduces sample diversity \citep{chen2026data,duan2026uncertainty,shaul2026parallel} and relies on a large bidirectional teacher that does not share the causal structure of AR generation \citep{chao2026seriality}.

A natural alternative is FlowMap distillation, which trains the student to jump directly along the teacher's denoising trajectories \citep{song2023consistency,kim2024consistency,frans2025one,lai2025principles,boffi2026build,geng2026mean,chen2026pi}. FlowMap distillation distills the teacher's trajectories directly, preserves its sample diversity, and recently matches DMD in quality on bidirectional video models \citep{shaul2026parallel,zhang2026simplifying}. Recent works have applied FlowMap distillation to AR video models \citep{zheng2026causal,zhao2026causal}, yet the resulting students lag behind SF-DMD students and still rely on a final SF-DMD stage to be competitive.

We hypothesize that the gap between FlowMap distillation and SF-DMD for AR video stems from at least two factors. First, an AR teacher cannot recover from a drifted history when supervising student rollouts. During self-forcing, the student's generated histories gradually drift away from real videos. SF-DMD trains the entire rollout sequence jointly with a bidirectional teacher, directly penalizing sequence-level drift. In FlowMap distillation, however, the AR teacher is conditioned on the student's drifted history. Because the AR teacher was trained only on clean histories, it simply continues the drift rather than correcting it. A student trained to match this teacher therefore accumulates drift over long rollouts. To provide a corrective signal, we propose \emph{\methodone} (Section~\ref{sec:antidrift}), which trains an auxiliary model to mimic the student's drifted transitions and uses it as a negative branch of guidance \citep{karras2024guiding} to push the teacher's prediction opposite to the drift, mitigating rollout drift in the distilled student (Figure~\ref{fig:limitations}(a)).

Second, motion is hard for a FlowMap student to learn, often resulting in blurry or distorted moving objects (Figure~\ref{fig:limitations}(b)). Unlike DMD, which only matches output distributions, a FlowMap student must directly approximate the teacher's velocity field in a few long steps \citep{yin2024improved}. In AR video, we find that while the static background is already fixed by the preceding frames, motion patches exhibit a complex, rapidly changing velocity field that is difficult for a few-step student to learn (Section~\ref{sec:motion}). To make motion easier to learn, we propose \emph{\methodtwo}, which locates motion patches from the teacher's uncertainty and applies stronger skip-layer guidance \citep{hyung2025spatiotemporal} to simplify their velocity field.

Because both guidance methods modify only the teacher's velocity field, they can be applied directly to any FlowMap objective. We evaluate them with three representative FlowMap methods---\piflow \citep{chen2026pi}, MeanFlow \citep{geng2026mean}, and consistency distillation (CD) \citep{song2023consistency,zheng2026causal}---using 1.3B AR video diffusion models \citep{wan2025wan,zhu2026causal,zheng2026causal}. Without any DMD stage, our recipe yields competitive 4-step FlowMap students across all three objectives while preserving the sample diversity of their AR teachers. In particular, our \piflow students distilled from 1.3B AR teachers match SF-DMD students distilled from 14B teachers on VBench. Our ablations further show that both guidance methods improve generation quality and our recipe reduces drift over 30\,s rollouts.
\section{Related Work}
\label{sec:related}

\begin{figure*}[t]
\centering
\includegraphics[width=0.65\linewidth]{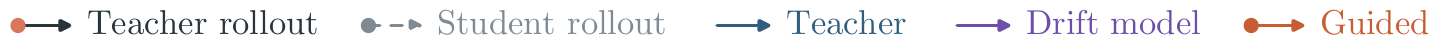}\\[2pt]
\begin{subfigure}[t]{0.30\linewidth}
\centering
\includegraphics[width=\linewidth]{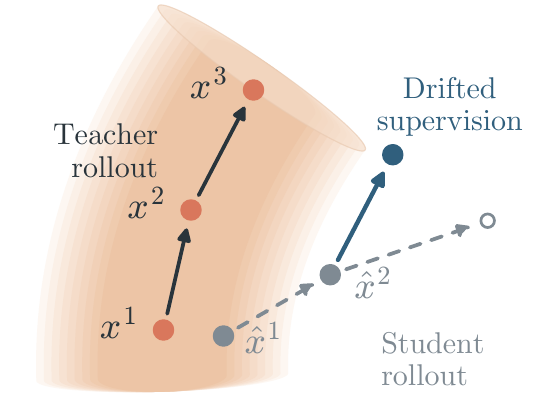}
\caption{Teacher on rollout}
\label{fig:teacher_on_rollout}
\end{subfigure}\hfill
\begin{subfigure}[t]{0.30\linewidth}
\centering
\includegraphics[width=\linewidth]{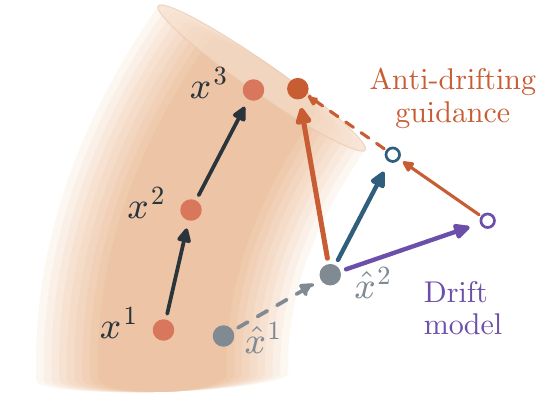}
\caption{\Methodone}
\label{fig:antidrift_panel}
\end{subfigure}\hfill
\begin{subfigure}[t]{0.37\linewidth}
\centering
\includegraphics[width=\linewidth]{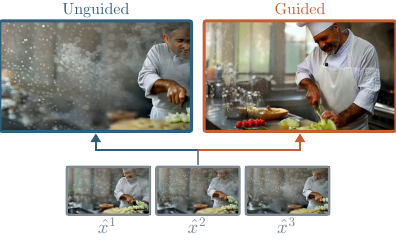}
\caption{Next chunk from a drifted rollout}
\label{fig:drift_teacher}
\end{subfigure}
\caption{\textbf{\Methodone.} (a)~The student rollout $(\hat{x}^1,\hat{x}^2,\dots)$ drifts away from the teacher rollout $(x^1,x^2,\dots)$ due to error accumulation. Conditioned on this drifted rollout, the AR teacher fails to correct the drift. (b)~\Methodone pushes the teacher's prediction away from a drift model that mimics the student's drifted transition, steering it back toward the teacher distribution. (c)~Next chunk generated from a drifted rollout without (Unguided) and with (Guided) \methodone.}
\label{fig:antidrift_guidance}
\vspace{-8pt}
\end{figure*}

\paragraph{Few-step AR video diffusion}
AR video diffusion models generate video chunks sequentially using causal attention and a KV cache \citep{chen2024diffusion,yin2025slow,teng2025magi}. When distilled into few-step students, these models suffer from exposure bias as errors accumulate across chunks. Most recent methods reduce exposure bias through self-forcing, training the student on its own rollouts with DMD \citep{huang2026self,zhu2026causal,zhao2026causal,zheng2026causal} or an adversarial loss \citep{lin2026autoregressive}. Orthogonally, recent works address KV-cache drift beyond the training horizon through KV recaching, attention sinks, and cache compression \citep{liu2026rolling,yang2025longlive,yi2025deep}, which are complementary to our distillation approach.

\paragraph{FlowMap distillation}
Consistency models map any noisy state along an ODE trajectory to its clean endpoint \citep{song2023consistency}, training either in discrete time \citep{luo2023latent,wang2024phased} or in continuous time with Jacobian--vector products \citep{lu2025simplifying,zheng2026large}. General FlowMap distillation extends this endpoint map to transitions between any two noise levels \citep{kim2024consistency,boffi2026build,sabour2026align,frans2025one,zhou2026terminal}. In particular, MeanFlow \citep{geng2026mean} predicts the average velocity over an interval, while \piflow \citep{chen2026pi,shaul2026parallel} predicts a velocity policy to match the teacher. For AR video generation, recent methods apply FlowMap distillation only as a teacher-forced initialization or an auxiliary regularizer for SF-DMD \citep{zheng2026causal,zhao2026causal}. In contrast, we use FlowMap distillation in both teacher-forcing and self-forcing without any DMD stage.

\paragraph{Guidance}
Classifier-free guidance (CFG) extrapolates a conditional prediction away from an unconditional negative branch \citep{ho2022classifier}. Subsequent methods construct the negative branch without dropping the condition by perturbing self-attention \citep{ahn2024self}, skipping network blocks \citep{hyung2025spatiotemporal}, or querying a weaker model \citep{karras2024guiding}. To reduce drift during multi-step AR sampling, prior work guides generation with altered histories \citep{song2025history,liao2026vpg} or compresses context frames \citep{zhang2025packing}. Most closely related to our \methodone, \citet{alemohammad2024self} fine-tune a copy of a diffusion model on its own samples to serve as a negative guide at inference. We instead train an auxiliary model online on the student's rollouts during self-forcing to distill an anti-drifting signal into a few-step student. 

\section{Preliminaries}
\label{sec:preliminaries}

\paragraph{AR video diffusion}
We represent a clean video latent as a sequence of $L$ chunks, $x_0=(x^1_0,\dots,x^L_0)$. An AR video diffusion model generates each chunk sequentially, conditioning chunk $i$ on its preceding history $x^{<i}_0$ through a causal KV cache. During training, flow matching \citep{lipman2022flow,liu2022flow,zhu2026causal} interpolates between the clean chunk $x^i_0$ and Gaussian noise $x^i_1\sim\mathcal{N}(0,I)$ at noise level $t\in[0,1]$ as $x^i_t=(1-t)x^i_0+tx^i_1$. The AR teacher $v_\phi(x^i_t,t\mid x^{<i}_0)$ is trained to predict the target velocity $x^i_1-x^i_0$.

\paragraph{Guidance}
Guidance extrapolates the teacher's conditional velocity $v_\phi$ away from a weaker negative branch $v_\phi^-$:
\begin{equation*}
    v^\star = v_\phi + (w-1)\,(v_\phi - v_\phi^-),
\end{equation*}
where $w>1$ is the guidance scale and we omit $(x^i_t,t\mid x^{<i}_0)$ for brevity. In standard CFG \citep{ho2022classifier}, $v_\phi$ and $v_\phi^-$ evaluate the teacher with the positive and negative text prompts, respectively. Using $v^\star$ as the distillation target transfers this guidance into the student, so the student requires no guidance at inference.

\paragraph{AR FlowMap distillation}
Given a history $x^{<i}_0$ and noise levels $0\le t<s\le 1$, the teacher's flow map transports chunk $i$ from $s$ to $t$ along the probability-flow ODE:
\begin{equation*}
    \Phi_{t\leftarrow s}(x^i_s\mid x^{<i}_0)=x^i_s-\int_t^s v^\star(x^i_\tau,\tau\mid x^{<i}_0)\,d\tau .
\end{equation*}
FlowMap distillation trains a student map $\Phi^\theta_{t\leftarrow s}$ to approximate $\Phi_{t\leftarrow s}$ directly so that each chunk can be generated in a few steps. In this work, we consider two families of objectives \citep{boffi2026build}. Lagrangian self-distillation (LSD), such as \piflow \citep{chen2026pi,shaul2026parallel}, matches the target-time derivative of $\Phi^\theta_{t\leftarrow s}$ to $v^\star$, while Eulerian self-distillation (ESD), such as MeanFlow \citep{geng2026mean} and consistency distillation (CD) \citep{song2023consistency,zheng2026causal}, enforces self-consistency across source times using $v^\star$. In both families, the student is supervised through the guided teacher velocity $v^\star(x^i_t,t\mid x^{<i}_0)$. For detailed formulations, see Appendix~\ref{sec:suppl_objectives}.

\paragraph{Two-stage AR distillation}
To train a student to generate each chunk $\hat{x}^i_0$ in a few steps, AR distillation typically proceeds in two stages that differ in how the history condition is formed \citep{huang2026self,zhu2026causal}. In Stage~1 (\emph{teacher-forcing}), each chunk is conditioned on the clean history $x^{<i}_0$ from the training data to provide a stable few-step initialization. In Stage~2 (\emph{self-forcing}) \citep{huang2026self}, each chunk is conditioned on the student's own rollout history $\hat{x}^{<i}_0$ to reduce exposure bias at inference. For an illustration, see Appendix~\ref{sec:suppl_forcing}.

\section{Method}
\label{sec:method}

We first analyze the two failure modes of standalone AR FlowMap distillation (Section~\ref{sec:analysis}). Based on this analysis, we introduce \emph{\methodone} to counter rollout drift during self-forcing (Section~\ref{sec:antidrift}) and \emph{\methodtwo} to make the teacher's velocity field easier for the student to learn, especially for motion (Section~\ref{sec:motion}). Finally, Section~\ref{sec:two_stage_recipe} combines both guidance methods into a two-stage AR FlowMap distillation recipe.

\subsection{Why AR FlowMap Distillation Fails}
\label{sec:analysis}

\begin{figure}[t]
\centering
\includegraphics[width=\linewidth]{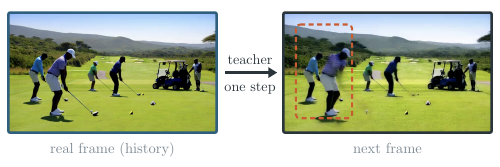}
\caption{\textbf{One-step prediction of the AR teacher.} Given a clean history frame, predicting the next frame in one step from noise keeps the static background sharp while blurring the moving player.}
\label{fig:one_nfe}
\vspace{-8pt}
\end{figure}

\paragraph{Rollout drift}
Existing AR distillation recipes already use FlowMap or ODE distillation during teacher-forcing \citep{zhu2026causal,zheng2026causal,zhao2026causal}, yet must switch to DMD during self-forcing because FlowMap distillation fails to prevent rollout drift. A student trained only on clean histories $x^{<i}_0$ suffers from exposure bias at inference time, so self-forcing trains the student on its own rollout histories $\hat{x}^{<i}_0$. In SF-DMD, a bidirectional teacher scores the entire generated sequence jointly, which directly penalizes sequence-level drift. In AR FlowMap distillation, however, the student is supervised by the AR teacher velocity $v_\phi(x^i_t,t\mid \hat{x}^{<i}_0)$. Because $v_\phi$ is also conditioned on the drifted history $\hat{x}^{<i}_0$, its prediction continues with the existing error rather than correcting it (Figure~\ref{fig:antidrift_guidance}(a,c), Unguided). Matching the teacher's velocity on student rollouts therefore preserves accumulated error rather than pulling the rollout back toward the teacher distribution.

\paragraph{Complex velocity field on motion patches}
In our preliminary experiments, distilling a few-step AR FlowMap student often produces blurry or distorted results on dynamic scenes, as seen in the bottom row of Figure~\ref{fig:limitations}(b). To see where few-step generation breaks down, Figure~\ref{fig:one_nfe} inspects a single Euler step of the AR teacher itself from a clean history and shows that one long step already reconstructs the static background sharply, yet leaves the moving player blurry, confirming that the failure concentrates on motion. Why is motion hard to learn in a few steps? Within each chunk, a FlowMap student must approximate the teacher's velocity field across long steps, and the learnability of this mapping is governed by the Lipschitz constant of the teacher's velocity field (see Proposition~\ref{prop:sensitivity} in Appendix~\ref{sec:suppl_prop}). After the first chunk, the preceding frames already fix the static background, whereas moving objects remain uncertain and induce a rapidly varying velocity field with a large Lipschitz constant on motion patches, making motion difficult for a few-step student to learn. 

\subsection{\Methodone}
\label{sec:antidrift}

To correct rollout drift during self-forcing, our key idea is to push the teacher's prediction in the opposite direction of the student's drift. To capture the direction in which the student drifts, we train an auxiliary model $v_\psi$---which we call the \emph{drift model}---to mimic the student's drifted transitions. Specifically, $v_\psi$ is initialized from the AR teacher $v_\phi$ and trained online via flow matching on the student's generated chunks $\hat{x}^i_0$, conditioned on its rollout histories $\hat{x}^{<i}_0$, with $n$ steps per student update. Using the teacher $v_\phi$ as the positive branch and the drift model $v_\psi$ as the negative branch, we guide the velocity as
\begin{equation*}
    v^\star = v_\phi + (\gamma - 1)\,(v_\phi - v_\psi),
\end{equation*}
where $v_\phi$ and $v_\psi$ are both evaluated at $(x^i_t,t\mid \hat{x}^{<i}_0)$, and $\gamma>1$ is the guidance scale. Because the student drifts further away from the teacher distribution than the teacher, the difference $v_\phi - v_\psi$ points opposite to the drift, steering the guided trajectory back toward the teacher distribution (Figure~\ref{fig:antidrift_guidance}(b,c), Guided).

\paragraph{Difference from DMD}
Although \methodone and DMD appear similar in contrasting the teacher with a model trained on student rollouts, they are fundamentally different. Empirically, whereas DMD is known to reduce sample diversity \citep{chen2026data,duan2026uncertainty,shaul2026parallel}, \methodone preserves or increases it in our experiments (Table~\ref{tab:abl_gamma}). This contrast comes from two key differences. First, DMD minimizes a reverse-KL divergence at the clean output $x_0$ by re-noising the student's generated video and backpropagating the score difference from a bidirectional teacher, whereas \methodone applies a reverse-KL correction at the intermediate noisy state $x^i_t$ along the denoising trajectory. Second, rather than optimizing a distribution-matching loss directly on the student, our correction acts only to tilt the target distribution of AR FlowMap distillation through the guided velocity $v^\star$, preserving the trajectory coupling between noise and output.

\subsection{\Methodtwo}
\label{sec:motion}

\begin{figure}[t]
\centering
\includegraphics[width=\linewidth]{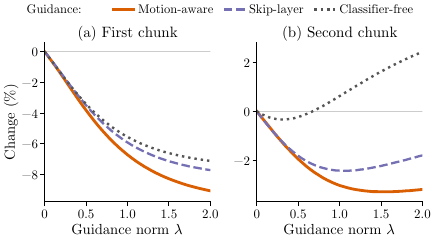}
\caption{\textbf{Normalized Lipschitz ratio of the guided teacher velocity.} Lower values indicate a smoother velocity field relative to the unguided teacher ($\lambda=0$). (a)~In the first chunk, all three guidance methods smooth the teacher's velocity field. (b)~In the second chunk, classifier-free guidance fails to smooth the velocity field and skip-layer guidance helps moderately, whereas \methodtwo achieves the largest reduction.}
\label{fig:curvature}
\vspace{-8pt}
\end{figure}

Next, we turn to making motion easier for a few-step student to learn. While investigating the learnability of the teacher's velocity field, we found that it depends closely on guidance. Figure~\ref{fig:curvature} measures the normalized Lipschitz ratio of the guided teacher velocity field, where lower values indicate a smoother, easier-to-learn velocity field (for the metric definition, see Appendix~\ref{sec:suppl_cv}). In the first chunk, standard CFG steadily lowers this ratio and smooths the velocity field. In the second chunk, however, CFG provides almost no smoothing effect because generation conditioned on the first chunk mainly involves frame-to-frame motion, which strengthening the text prompt fails to guide effectively. By contrast, skip-layer guidance (SLG) \citep{hyung2025spatiotemporal}, an unconditional guidance method that constructs the negative branch by skipping a teacher block without relying on the text prompt, smooths the velocity field even in the second chunk.

\begin{figure}[t]
\centering
\input{figures/uncertainty}
\vspace{4pt}
\includegraphics[width=\linewidth]{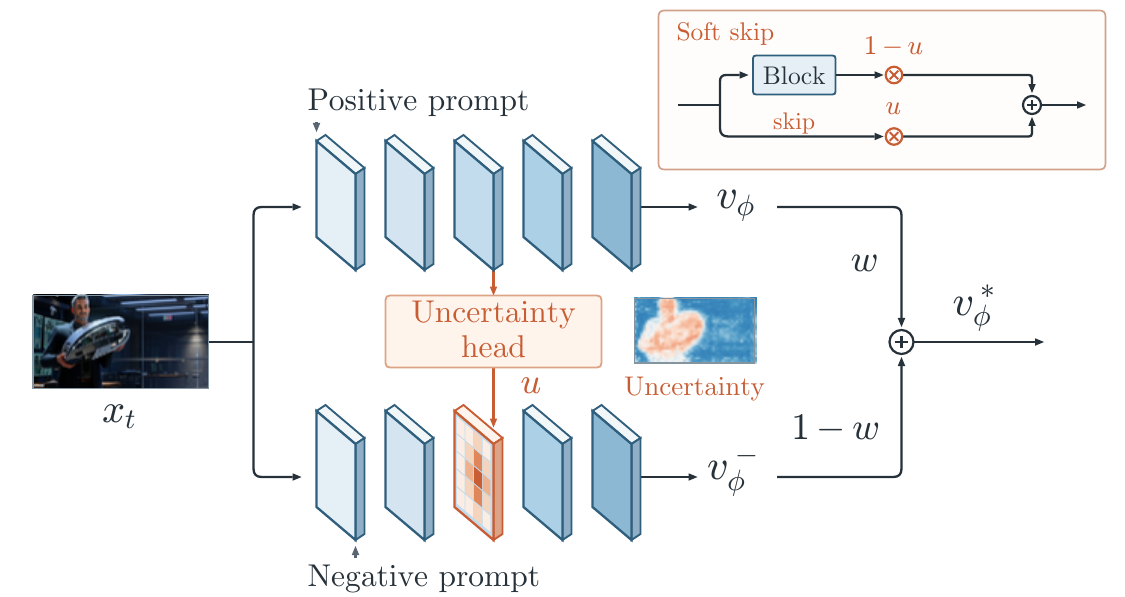}
\caption{\textbf{\Methodtwo.} Stronger skip-layer guidance is applied to motion patches identified by the teacher's uncertainty map. An uncertainty head on the positive branch $v_\phi$ modulates the soft-skip weight on the negative branch $v_\phi^-$ in Eq.~\eqref{eq:soft_skip}.}
\label{fig:motion_guidance}
\vspace{-8pt}
\end{figure}

\paragraph{Uncertainty-weighted soft skip}
Motivated by this finding, we tailor SLG to AR video distillation. Because a FlowMap student mainly struggles with motion in later chunks, our idea is to apply SLG selectively to motion patches while avoiding unnecessary guidance on static patches. Since the preceding frames already fix the static background, the teacher's prediction uncertainty naturally locates motion patches, as shown in Figure~\ref{fig:uncertainty}. Based on this observation, \emph{\methodtwo} applies stronger skip-layer guidance to patches with higher teacher uncertainty (Figure~\ref{fig:motion_guidance}). Specifically, we attach a lightweight head to block $\ell$ of the teacher to predict each patch's uncertainty $u$. When evaluating the negative branch $v_\phi^-$, we use $u$ to apply a patch-wise soft skip at block $\ell$:
\begin{equation}
    z = z^{\mathrm{no\text{-}skip}} + u \odot (z^{\mathrm{skip}} - z^{\mathrm{no\text{-}skip}}),
    \label{eq:soft_skip}
\end{equation}
where $z^{\mathrm{no\text{-}skip}}$ and $z^{\mathrm{skip}}$ are the block's output and input features. Static patches with $u\approx 0$ thus receive only standard CFG, whereas motion patches with large $u$ receive stronger skip-layer guidance. As shown in Figure~\ref{fig:curvature}(b), this selective guidance makes the teacher's velocity field substantially smoother in later chunks. The uncertainty head is trained beforehand on the teacher's flow-matching residual with a Gaussian negative log-likelihood loss \citep{kendall2017uncertainties}, and $u$ is normalized during distillation (for details, see Appendix~\ref{sec:suppl_uncertainty}).

\begin{table*}[t]
\centering
\caption{\textbf{4-NFE AR students on Wan2.1-1.3B.} AR base: the AR model each student is initialized from; rows are grouped by it. Type: final-stage objective. Teacher: distillation teacher (AR: causal teacher; others: bidirectional). $^\dagger$Official checkpoint. Within each group, {\setlength{\fboxsep}{1pt}\colorbox{vbbestbg}{\textbf{bold}}: best; \colorbox{vbsecondbg}{\underline{underlined}}: second; \colorbox{vbthirdbg}{shaded}: third} (AR teachers in \textcolor{rankref}{gray} for reference).}
\label{tab:main}
\footnotesize
\setlength{\tabcolsep}{3pt}
\begin{tabular}{llllccccccccc}
\toprule
& & & & & \multicolumn{4}{c}{VBench $\uparrow$} & \multicolumn{2}{c}{V-JEPA 2 $\uparrow$} & \multicolumn{2}{c}{VideoMAE V2 $\uparrow$} \\
\cmidrule(lr){6-9}\cmidrule(lr){10-11}\cmidrule(lr){12-13}
AR base & Method & Type & Teacher & NFE & Total & Quality & Semantic & Dynamic & Cos & L2 & Cos & L2 \\
\midrule
\citep{huang2026self} & Self Forcing$^\dagger$ & DMD & 14B & 4 & 84.22 & 84.83 & 81.81 & 66.1 & 0.0555 & 21.5 & 0.0117 & 2.50 \\ 
\cmidrule{1-13}
\multirow{6}{*}{\citep{zheng2026causal}} & \unranked AR teacher, CFG & \unranked -- & \unranked -- & \unranked $32\times2$ & \unranked 84.38 & \unranked 84.85 & \unranked 82.47 & \unranked 65.0 & \unranked 0.0624 & \unranked 23.2 & \unranked 0.0128 & \unranked 2.60 \\ 
 & Causal-rCM$^\dagger$ & DMD & 14B\textsuperscript{AR}, 14B & 4 & \vbbest{84.92} & \vbsecond{85.44} & \vbbest{82.82} & \vbbest{81.1} & \vbthird{0.0666} & 23.6 & \vbsecond{0.0166} & \vbsecond{2.95} \\ 
 & Causal-rCM, TF-sCM$^\dagger$ & FlowMap & 14B\textsuperscript{AR} & 4 & 83.78 & 84.49 & 80.92 & 67.5 & \vbbest{0.0789} & \vbbest{26.2} & \vbbest{0.0182} & \vbbest{3.13} \\ 
 & Ours (\piflow) & FlowMap & 1.3B\textsuperscript{AR} & 4 & \vbsecond{84.87} & \vbbest{85.76} & 81.29 & \vbsecond{71.4} & \vbsecond{0.0735} & \vbsecond{25.2} & \vbthird{0.0143} & \vbthird{2.77} \\ 
 & Ours (MeanFlow) & FlowMap & 1.3B\textsuperscript{AR} & 4 & \vbthird{84.66} & 85.19 & \vbsecond{82.55} & 65.8 & 0.0656 & \vbthird{23.7} & 0.0126 & 2.58 \\ 
 & Ours (CD) & FlowMap & 1.3B\textsuperscript{AR} & 4 & 84.54 & \vbthird{85.20} & \vbthird{81.91} & \vbthird{68.3} & 0.0634 & 23.4 & 0.0122 & 2.56 \\ 
\cmidrule{1-13}
\multirow{5}{*}{\citep{zhu2026causal}} & \unranked AR teacher, CFG & \unranked -- & \unranked -- & \unranked $32\times2$ & \unranked 84.46 & \unranked 85.11 & \unranked 81.86 & \unranked 77.8 & \unranked 0.0882 & \unranked 26.9 & \unranked 0.0236 & \unranked 3.50 \\ 
 & Causal Forcing$^\dagger$ & DMD & 14B & 4 & \vbsecond{84.74} & \vbsecond{85.41} & \vbbest{82.02} & \vbsecond{83.9} & 0.0745 & 24.4 & 0.0198 & 3.24 \\ 
 & Ours (\piflow) & FlowMap & 1.3B\textsuperscript{AR} & 4 & \vbbest{84.76} & \vbbest{85.83} & \vbthird{80.46} & \vbbest{88.3} & \vbbest{0.0964} & \vbbest{27.8} & \vbbest{0.0281} & \vbbest{3.81} \\ 
 & Ours (MeanFlow) & FlowMap & 1.3B\textsuperscript{AR} & 4 & \vbthird{84.20} & \vbthird{84.80} & \vbsecond{81.80} & \vbthird{69.7} & \vbthird{0.0842} & \vbthird{26.3} & \vbthird{0.0216} & \vbthird{3.32} \\ 
 & Ours (CD) & FlowMap & 1.3B\textsuperscript{AR} & 4 & 83.70 & 84.55 & 80.32 & \vbthird{69.7} & \vbsecond{0.0875} & \vbsecond{26.6} & \vbsecond{0.0241} & \vbsecond{3.54} \\ 
\bottomrule
\end{tabular}
\vspace{-8pt}
\end{table*}

\begin{figure*}[t]
\centering
\includegraphics[width=\linewidth]{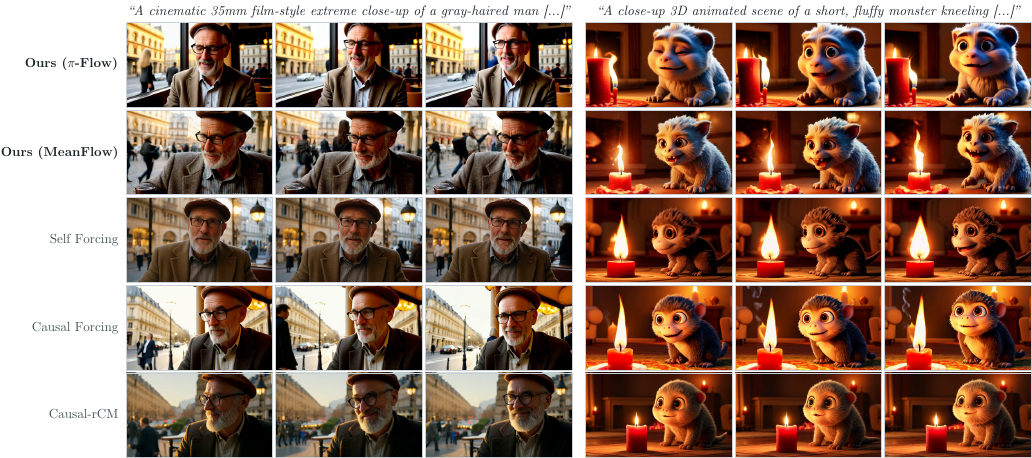}
\caption{\textbf{Qualitative comparison of generation quality.} Frames at $0.5$, $2.5$, and $4.5$\,s from the same prompt and seed per panel.}
\label{fig:\currfilebase}%
\vspace{-4pt}
\end{figure*}

\subsection{Two-Stage FlowMap Distillation}
\label{sec:two_stage_recipe}

Combining \methodtwo and \methodone gives the guided teacher velocity:
\begin{equation}
    v^\star=v_\phi+(w-1)\,(v_\phi-v_\phi^-)+(\gamma-1)\,(v_\phi-v_\psi),
    \label{eq:target}
\end{equation}
where $w$ and $\gamma$ are the scales for \methodtwo and \methodone. We plug $v^\star$ directly into the FlowMap distillation objective (such as \piflow, MeanFlow, or CD). In Stage~1 (teacher-forcing), the student is trained on clean histories with \methodtwo alone by setting $\gamma=1$. In Stage~2 (self-forcing), the student is trained on its own rollout histories, and we enable \methodone with $\gamma>1$ on all chunks after the first.

\section{Experiments}
\label{sec:experiments}

\subsection{Setup}
\label{sec:setup}

\paragraph{Teachers and guidance}
We distill the AR teachers released by Causal-rCM \citep{zheng2026causal} and Causal Forcing \citep{zhu2026causal}. Both are causal Wan2.1-T2V-1.3B models \citep{wan2025wan} trained with teacher-forcing on synthetic videos from Wan2.1-T2V-14B. We use their chunk-wise teachers, which generate 81-frame videos at $832\times480$ in seven chunks of three latent frames. To evaluate the teachers in Table~\ref{tab:main}, we sample them with 32 Euler steps ($32\times2=64$ NFE with CFG) and a timestep shift of 6. We set both the \methodone and \methodtwo scales to $\gamma=w=5$ in Eq.~\eqref{eq:target}, and apply the soft skip of Eq.~\eqref{eq:soft_skip} at block $\ell=10$. For implementation details on uncertainty prediction and normalization, see Appendix~\ref{sec:suppl_uncertainty}.

\paragraph{Distillation}
All students are initialized from their respective AR teachers and generate each chunk in 4 NFE without guidance at inference. Stage~1 trains the student with teacher-forcing on clean video latents from Wan2.1-T2V-14B \citep{wan2025wan}, whereas Stage~2 trains it with self-forcing on its own rollouts from the extended VidProM prompts \citep{wang2024vidprom,huang2026self}. In Stage~2, the drift model is initialized from the teacher and updated on the student's rollouts for $n=5$ steps per student update for \piflow and $n=1$ step for MeanFlow and CD. For training details and hyperparameters of \piflow, MeanFlow, and CD, see Appendix~\ref{sec:suppl_training}.

\begin{figure*}[t]
\centering
\includegraphics[width=\linewidth]{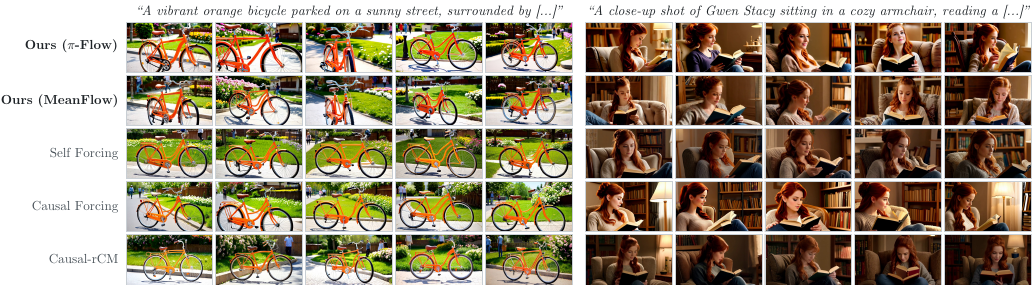}
\caption{\textbf{Qualitative comparison of sample diversity.} Frames at $2.5$\,s across five seeds for the same prompt.}
\label{fig:\currfilebase}%
\end{figure*}
\begin{figure*}[!t]
\centering
\includegraphics[width=\linewidth]{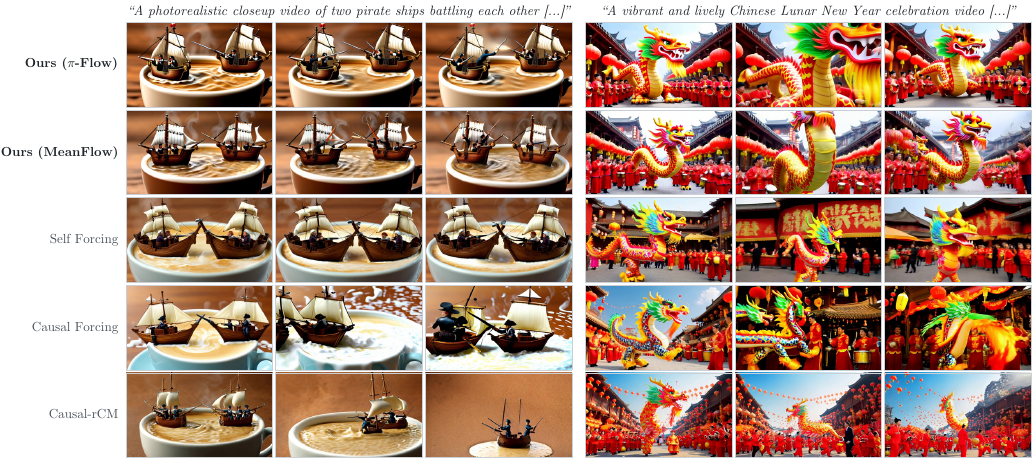}
\caption{\textbf{Qualitative comparison of 30\,s video generation.} Frames at $0$, $10$, and $30$\,s from the same prompt and seed per panel, using the KV-cache pattern of \citet{yi2025deep}.}
\label{fig:\currfilebase}%
\vspace{-4pt}
\end{figure*}

\paragraph{Baselines}
We compare our students with three SF-DMD baselines, Self Forcing \citep{huang2026self}, Causal-rCM \citep{zheng2026causal}, and Causal Forcing \citep{zhu2026causal}. All three use a 14B bidirectional teacher during SF-DMD, and Causal-rCM also uses a 14B AR teacher during Stage~1. We additionally include TF-sCM \citep{zheng2026causal}, the Stage~1 FlowMap student from Causal-rCM before SF-DMD. For a fair comparison, we pair each SF-DMD baseline with our students initialized from the same released AR model, since this model strongly affects the distilled student's quality and diversity. We evaluate all official checkpoints using their default inference settings.

\paragraph{Evaluation}
We evaluate 5\,s generation on the 946 extended prompts of Self Forcing \citep{huang2026self} over five seeds using VBench \citep{huang2024vbench} Total, Quality, Semantic, and Dynamic scores. Following PDD \citep{shaul2026parallel}, we measure sample diversity on the same five seeds as the mean pairwise feature distance among the five videos per prompt. We report cosine ($1-\cos$) and L2 distances in V-JEPA~2 \citep{assran2025v} and VideoMAE~V2 \citep{wang2023videomae} feature spaces. To evaluate long-horizon drift, we generate 30\,s videos on the 128 MovieGen prompts from Deep Forcing \citep{yi2025deep} using its KV-cache management and split each video into 16 clips of about 2\,s. We report $\Delta$IQ and $\Delta$AQ, the mean VBench imaging and aesthetic quality of the last three clips minus those of the first three, averaged over five seeds.

\subsection{Results}
\label{sec:main_results}

\paragraph{5\,s generation quality}
Table~\ref{tab:main} and Figure~\ref{fig:qualitative} compare the 4-NFE students on 5\,s video generation. Without any DMD stage, our \piflow students match the overall VBench scores of the SF-DMD baselines distilled from 14B teachers across both AR bases, while ranking first in VBench Quality in both groups. Our MeanFlow and CD students are also competitive, surpassing the Stage~1 FlowMap baseline TF-sCM on the Causal-rCM base. 

\paragraph{Sample diversity}
As shown in Table~\ref{tab:main} and Figure~\ref{fig:diversity}, because AR FlowMap distillation learns the teacher's trajectories directly rather than optimizing a reverse-KL DMD loss, our students preserve the sample diversity of their 64-NFE AR teachers.

\begin{table}[t]
\centering
\caption{\textbf{30\,s generation.} DD: dynamic degree; AQ / IQ: aesthetic / imaging quality; $\Delta$IQ / $\Delta$AQ: start-to-end quality change, using the KV-cache pattern of \citet{yi2025deep}. Colors follow Table~\ref{tab:main}.}
\label{tab:long}
\footnotesize
\setlength{\tabcolsep}{3pt}
\begin{tabularx}{0.9\columnwidth}{Xccccc}
\toprule
Method & DD $\uparrow$ & AQ $\uparrow$ & IQ $\uparrow$ & $\Delta$IQ $\uparrow$ & $\Delta$AQ $\uparrow$ \\
\midrule
Self Forcing$^{\dagger}$ & 60.8 & 61.18 & 69.30 & $-$0.76 & $-$2.00 \\ 
\midrule
\multicolumn{6}{l}{\textit{Causal-rCM teacher}} \\
Causal-rCM$^{\dagger}$ & \vbbest{65.6} & \vbthird{61.41} & 65.92 & $-$3.28 & \vbthird{$-$4.06} \\ 
Causal-rCM, TF-sCM$^{\dagger}$ & \vbsecond{48.6} & 58.80 & 65.29 & $-$3.10 & $-$5.08 \\ 
Ours (\piflow) & 41.1 & \vbbest{62.13} & \vbsecond{67.41} & \vbbest{$-$1.64} & \vbbest{$-$3.83} \\ 
Ours (MeanFlow) & \vbthird{41.7} & \vbsecond{61.81} & \vbthird{67.19} & \vbsecond{$-$1.95} & $-$4.12 \\ 
Ours (CD) & 41.3 & 60.27 & \vbbest{68.25} & \vbthird{$-$1.97} & \vbsecond{$-$4.01} \\ 
\midrule
\multicolumn{6}{l}{\textit{Causal Forcing teacher}} \\
Causal Forcing$^{\dagger}$ & \vbsecond{81.7} & \vbbest{58.71} & \vbbest{68.92} & \vbbest{$-$1.44} & \vbbest{$-$5.59} \\ 
Ours (\piflow) & \vbbest{82.1} & \vbthird{57.92} & \vbsecond{64.62} & \vbsecond{$-$4.94} & \vbsecond{$-$6.80} \\ 
Ours (MeanFlow) & \vbthird{59.8} & \vbsecond{58.37} & \vbthird{63.12} & \vbthird{$-$6.50} & \vbthird{$-$8.47} \\ 
Ours (CD) & 49.4 & 54.89 & 62.04 & $-$9.01 & $-$11.56 \\ 
\bottomrule
\end{tabularx}
\vspace{-6pt}
\end{table}

\paragraph{Long-video generation}
Table~\ref{tab:long} and Figure~\ref{fig:longvideo} report 30\,s video generation results. With the Causal-rCM teacher, all three of our FlowMap students exhibit less quality drift ($\Delta$IQ) and higher imaging quality than both Causal-rCM and TF-sCM, with \piflow achieving the smallest drift. With the Causal Forcing teacher, our students show larger drift than with the Causal-rCM teacher while achieving higher dynamic degree (DD).

\subsection{Ablation Studies}
\label{sec:ablation}

We conduct ablations on the Causal Forcing base \citep{zhu2026causal}. All students use 4 NFE per chunk, and VBench denotes the Total score averaged over five seeds.

\begin{table}[t]
\centering
\caption{\textbf{\Methodone does not hurt sample diversity.} \piflow students across $n$ drift-model updates per student update, where $\gamma=1$ disables \methodone.}
\label{tab:abl_gamma}
\footnotesize
\setlength{\tabcolsep}{4.5pt}
\begin{tabular}{ccccc}
\toprule
$n$ & $\gamma$ & VBench $\uparrow$ & V-JEPA 2 $\uparrow$ & VideoMAE V2 $\uparrow$ \\
\midrule
\rowcolor{gray!15} 5 & 5 (ours) & \textbf{84.76} & \textbf{0.0964} & \textbf{0.0281} \\
1 & 5 & 84.56 & 0.0902 & 0.0252 \\
-- & 1 & 84.35 & 0.0905 & 0.0239 \\
\bottomrule
\end{tabular}
\vspace{-6pt}
\end{table}

\paragraph{\Methodone}
Table~\ref{tab:abl_gamma} varies the number of drift-model updates $n$ per student update at $\gamma=5$. Applying \methodone and training the drift model more accurately with larger $n$ ($n=5$) improves both 5\,s VBench and sample diversity, confirming that \methodone does not hurt diversity (for Causal-rCM results, see Appendix~\ref{sec:suppl_drift}). On 30\,s rollouts, Table~\ref{tab:abl_stage2} compares training with and without \methodone. Across both objectives, \methodone reduces 30\,s quality drift ($\Delta$IQ and $\Delta$AQ) while preserving dynamic degree (DD), showing that it mitigates drift without sacrificing motion.

\begin{table}[t]
\centering
\caption{\textbf{\Methodone in Stage~2.} 30\,s rollout metrics with \methodtwo in both stages, where $\gamma=1$ disables \methodone.}
\label{tab:abl_stage2}
\footnotesize
\setlength{\tabcolsep}{3pt}
\begin{tabularx}{0.9\columnwidth}{lXccc}
\toprule
Method & Teacher guidance & DD $\uparrow$ & $\Delta$IQ $\uparrow$ & $\Delta$AQ $\uparrow$ \\
\midrule
\piflow & \cellcolor{gray!15}Anti-drifting (ours) & \cellcolor{gray!15}\textbf{82.1} & \cellcolor{gray!15}\textbf{$-$4.94} & \cellcolor{gray!15}\textbf{$-$6.80} \\
 & \quad $-$ \methodone & 62.7 & $-$7.79 & $-$9.49 \\
\midrule
CD & \cellcolor{gray!15}Anti-drifting (ours) & \cellcolor{gray!15}\textbf{49.4} & \cellcolor{gray!15}\textbf{$-$9.01} & \cellcolor{gray!15}\textbf{$-$11.56} \\
 & \quad $-$ \methodone & 39.5 & $-$10.00 & $-$16.51 \\
\bottomrule
\end{tabularx}
\end{table}

\begin{table}[t]
\centering
\caption{\textbf{\Methodtwo in Stage~1.} Stage~1 \piflow students, where Skip-layer applies uniform skip-layer guidance.}
\label{tab:abl_motion}
\footnotesize
\begin{tabularx}{0.9\columnwidth}{Xc}
\toprule
Teacher guidance & VBench $\uparrow$ \\
\midrule
\rowcolor{gray!15} Motion-aware (ours) & \textbf{84.01} \\ 
Skip-layer & 83.43 \\ 
Classifier-free & 83.66 \\ 
\bottomrule
\end{tabularx}
\end{table}

\begin{table}[t]
\centering
\caption{\textbf{Contribution of each guidance method.} Stage~2 \piflow students, where removing both leaves CFG alone in both stages.}
\label{tab:abl_components}
\footnotesize
\begin{tabularx}{0.9\columnwidth}{Xc}
\toprule
Variant & VBench $\uparrow$ \\
\midrule
\rowcolor{gray!15} Ours & \textbf{84.76} \\ 
\quad $-$ \methodone & 84.35 \\ 
\quad $-$ \methodtwo & 84.19 \\ 
\quad $-$ both (CFG only) & 81.61 \\ 
\bottomrule
\end{tabularx}
\end{table}

\paragraph{\Methodtwo}
Table~\ref{tab:abl_motion} compares \methodtwo with uniform skip-layer guidance and CFG alone on Stage~1 \piflow students. Uniform skip-layer guidance degrades quality relative to CFG, whereas \methodtwo improves over both by selectively guiding uncertain motion patches (for results across fixed skip weights and CD, see Table~\ref{tab:abl_stage1} in Appendix~\ref{sec:suppl_stage1}).

\paragraph{Components}
Table~\ref{tab:abl_components} ablates \methodone and \methodtwo on our \piflow student. Removing both guidance methods substantially degrades 5\,s VBench, adding either method alone recovers most of the gap, and combining both achieves the best quality, showing that both guidance methods are important and complementary. 

\section{Conclusion and Limitations}
\label{sec:conclusion}

We presented a two-stage FlowMap distillation recipe for autoregressive video models that requires no DMD stage. Our \emph{\methodone} and \emph{\methodtwo} modify only the teacher's velocity field, so the same recipe applies to \piflow, MeanFlow, and consistency distillation. Our results demonstrate that 4-step \piflow students distilled from 1.3B AR teachers match SF-DMD students distilled from 14B teachers on VBench, while preserving the sample diversity of their AR teachers and mitigating drift over 30\,s rollouts.

\paragraph{Limitations.}
Because FlowMap distillation follows the AR teacher's trajectories, our students inherit the teacher's tendency to drift over long rollouts. Our \methodone mitigates but does not fully eliminate this drift, and combining it with stronger teachers or longer-horizon supervision remains an open direction.

\paragraph{Acknowledgments.}
We would like to thank Aryan V~S, Donghoon Ahn, Sangyoon Lee, and Susung Hong for helpful discussions.

{
    \small
    \bibliographystyle{ieeenat_fullname}
    \bibliography{references}
}

\clearpage
\maketitlesupplementary
\appendix

\section{Additional Limitations}
\label{sec:suppl_limitations}

Beyond rollout drift, our students inherit other properties of the 1.3B AR teachers. While they achieve the highest VBench Quality scores, they inherit lower Semantic scores than SF-DMD students distilled from 14B teachers, as well as the lower dynamic degree of the 1.3B Causal-rCM teacher (Table~\ref{tab:main}). Maintaining the online drift model also requires $n$ additional updates per student update during Stage~2.

\section{Why Motion Is Hard to Distill}
\label{sec:suppl_analysis}

This section connects why a few-step student struggles to learn motion to the sensitivity of the teacher's velocity field. Sufficiently high posterior uncertainty induces a large velocity Jacobian norm, removing smoothness guarantees on the flow map (Section~\ref{sec:analysis}). Empirically, we measure a scale-normalized finite-difference ratio $C_v$ and show that \methodtwo lowers $C_v$ on uncertain motion patches (Section~\ref{sec:motion}).

\subsection{Uncertainty and the Lipschitz Constant}
\label{sec:suppl_prop}

We first show that sufficiently large posterior uncertainty implies a large velocity Jacobian norm. For brevity, we omit the chunk index $i$ and the history $x^{<i}_0$. Let $x_t=(1-t)x_0+tx_1$ with $x_1\sim\mathcal{N}(0,I)$ independent of $x_0$, let $v(x_t,t)=\bigl(x_t-\mathbb{E}[x_0\mid x_t]\bigr)/t$ be the Bayes-optimal velocity, and let $\Sigma_t(x_t)=\mathrm{Cov}(x_0\mid x_t)$ be the posterior covariance of the clean chunk, which measures how uncertain the teacher is about it.

\begin{proposition}[Uncertainty and the Lipschitz constant]
\label{prop:sensitivity}
(i)~Sufficiently large posterior uncertainty implies a large velocity Jacobian norm:
\begin{equation*}
\begin{gathered}
    \nabla_{x_t}v=\frac{1}{t}\Bigl(I-\frac{1-t}{t^2}\Sigma_t\Bigr),\\
    \|\nabla_{x_t}v\|_{\mathrm{op}}\ge\frac{1-t}{t^3}\|\Sigma_t\|_{\mathrm{op}}-\frac{1}{t}.
\end{gathered}
\end{equation*}
(ii)~The flow-matching residual $r=v(x_t,t)-(x_1-x_0)$ measures uncertainty:
\begin{equation*}
    \mathbb{E}\bigl[\|r\|^2\mid x_t\bigr]=\frac{1}{t^2}\,\mathrm{tr}\,\Sigma_t.
\end{equation*}
(iii)~The Lipschitz constant bounds the flow map: for any velocity field with Lipschitz constant $L_\tau=\sup_x\|\nabla_x v(x,\tau)\|_{\mathrm{op}}$,
\begin{equation*}
    \|\Phi_{t\leftarrow s}(x_a)-\Phi_{t\leftarrow s}(x_b)\|\le\exp\Bigl(\int_t^s L_\tau\,d\tau\Bigr)\|x_a-x_b\|.
\end{equation*}
\end{proposition}
\begin{proof}
\textit{(i)}~Since $p(x_t\mid x_0)=\mathcal{N}\bigl((1-t)x_0,t^2I\bigr)$, its score is
\begin{equation*}
    \nabla_{x_t}\log p(x_t\mid x_0)=\frac{(1-t)x_0-x_t}{t^2}.
\end{equation*}
Differentiating $\mathbb{E}[x_0\mid x_t]=\int x_0\,p(x_0\mid x_t)\,dx_0$ with respect to $x_t$ therefore gives
\begin{equation*}
    \nabla_{x_t}\mathbb{E}[x_0\mid x_t]=\frac{1-t}{t^2}\,\Sigma_t.
\end{equation*}
Substituting this into $v=\bigl(x_t-\mathbb{E}[x_0\mid x_t]\bigr)/t$ gives the expression for $\nabla_{x_t}v$, and the bound follows from $\|A-B\|_{\mathrm{op}}\ge\|B\|_{\mathrm{op}}-\|A\|_{\mathrm{op}}$.

\textit{(ii)}~Substituting $x_1-x_0=(x_t-x_0)/t$ into the residual gives
\begin{equation*}
    r=\frac{x_0-\mathbb{E}[x_0\mid x_t]}{t},
\end{equation*}
whose conditional second moment is $\mathrm{tr}\,\Sigma_t/t^2$.

\textit{(iii)}~For two trajectories $x_\tau=\Phi_{\tau\leftarrow s}(x_a)$ and $y_\tau=\Phi_{\tau\leftarrow s}(x_b)$, the distance $\delta(\tau)=\|x_\tau-y_\tau\|$ satisfies
\begin{equation*}
    \delta(t)\le\delta(s)+\int_t^s L_\tau\,\delta(\tau)\,d\tau,
\end{equation*}
so Gr\"onwall's inequality gives $\delta(t)\le\delta(s)\exp\bigl(\int_t^s L_\tau\,d\tau\bigr)$.
\end{proof}

Part~(i) shows that once posterior uncertainty is sufficiently large ($\|\Sigma_t\|_{\mathrm{op}} > \frac{2t^2}{1-t}$), higher uncertainty forces the velocity field to vary sharply with the input. Part~(iii) says that a velocity field with a small Lipschitz constant gives a smooth flow map, which a few-step student can fit with long jumps, while a large Lipschitz constant removes this guarantee. Part~(ii) shows that the flow-matching residual measures this uncertainty, which is why we train the uncertainty head on it (Section~\ref{sec:suppl_uncertainty}). In AR video, the history already fixes static regions, so the uncertainty concentrates on moving objects.

\subsection{Normalized Lipschitz Ratio}
\label{sec:suppl_cv}

To evaluate how smoothly the guided teacher velocity varies across nearby states while controlling for changes in overall velocity magnitude $\|v^\star\|$ across guidance strengths, we measure the scale-normalized finite-difference Lipschitz ratio between two nearby noisy states $x_a$ and $x_b$ at the same noise level:
\begin{equation}
    C_v=\frac{\|v^\star(x_a)-v^\star(x_b)\|\,/\,\|v^\star\|}{\|x_a-x_b\|\,/\,\|x\|},
    \label{eq:cv}
\end{equation}
where $\|\cdot\|$ is the root-mean-square norm, and $\|v^\star\|$ and $\|x\|$ normalize the velocity and state magnitudes over the pair. Unlike the exact Lipschitz constant $\sup_x\|\nabla_x v^\star(x,t)\|_{\mathrm{op}}$, $C_v$ is evaluated on finite-difference pairs and normalized by $\|v^\star\|$, so it is invariant to uniform rescaling of $v^\star$; a lower $C_v$ therefore indicates reduced relative variation across nearby states rather than a guaranteed reduction in the unnormalized Lipschitz constant.

Because the root-mean-square magnitude of the guidance direction $g=v_\phi-v_\phi^-$ differs across guidance methods at the same scale $w$, we compare directions at matched guidance magnitude by normalizing $g$ at each noisy state and parameterizing the guided velocity by its RMS guidance norm $\lambda\ge 0$:
\begin{equation*}
    v^\star_\lambda = v_\phi + \lambda\,\frac{g}{\|g\|},
\end{equation*}
where standard guidance at scale $w$ corresponds to $\lambda=(w-1)\|g\|$, and $\lambda=0$ recovers the unguided teacher $v_\phi$. Figure~\ref{fig:curvature} sweeps $\lambda\in[0,2]$ over whole chunks (with uniform skip-layer guidance using each state's mean skip weight $u_p$) and plots the percentage change $100\times\bigl(\bar{C}_v(\lambda)/\bar{C}_v(0)-1\bigr)$ of the clip- and time-averaged ratio $\bar{C}_v(\lambda)$ relative to $\lambda=0$. For Figure~\ref{fig:suppl_deciles}, we compute the patch-wise ratio $C_v^{\mathrm{local}}$ under small random perturbations across ten uncertainty deciles $D_1,\dots,D_{10}$ in the second chunk, matching $\lambda=(w-1)\|g_{\mathrm{CFG}}\|$ at $w=5$ for all methods; panel~(a) shows $\bar{C}_v^{\mathrm{local}}$, which rises sharply on the most uncertain patches under CFG, and panel~(b) reports the percentage gain in $1/\bar{C}_v^{\mathrm{local}}$ relative to CFG, $100\times\bigl(\bar{C}_{v,\mathrm{CFG}}^{\mathrm{local}}/\bar{C}_v^{\mathrm{local}}-1\bigr)$.

\subsection{Guidance and the Normalized Lipschitz Ratio}
\label{sec:suppl_deciles}

\begin{figure}[t]
\centering
\includegraphics[width=\linewidth]{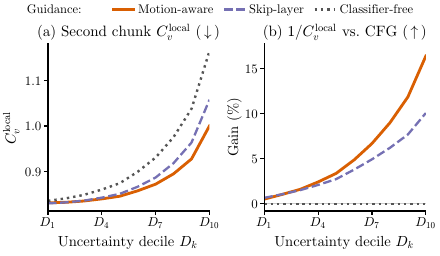}
\caption{\textbf{Normalized Lipschitz ratio across uncertainty deciles in the second chunk.} Patches are grouped into deciles $D_1,\dots,D_{10}$ by predicted uncertainty, and all guidance methods use the same guidance norm. (a)~With classifier-free guidance, $C_v^{\mathrm{local}}$ rises sharply on the most uncertain patches. (b)~Relative to classifier-free guidance, \methodtwo lowers $C_v^{\mathrm{local}}$ more than uniform skip-layer guidance, especially on the most uncertain patches.}
\label{fig:suppl_deciles}
\end{figure}

Guidance directly affects this normalized Lipschitz ratio. As shown in Figure~\ref{fig:curvature}, stronger CFG steadily lowers $C_v$ in the first chunk, making the velocity field smoother, whereas in the second chunk it does not. We attribute this to the preceding frames: once the chunk is conditioned on them, the text prompt has less influence on what remains to be generated, so strengthening it does little to smooth the velocity field. While skip-layer guidance \citep{hyung2025spatiotemporal} perturbs the model by skipping a block rather than relying on text contrast, in our framework we combine it with CFG: both uniform skip-layer guidance and \methodtwo apply soft skips (Eq.~\eqref{eq:soft_skip}) to the negative-prompt branch $v_\phi^-$ of CFG (Figure~\ref{fig:motion_guidance}), recovering standard CFG when $u=0$. Skipping a block on $v_\phi^-$ lowers $C_v$ even in the second chunk, and \methodtwo lowers $C_v^{\mathrm{local}}$ most on the most uncertain patches, outperforming uniform skip-layer guidance at the same guidance norm (Figure~\ref{fig:suppl_deciles}(b)).

\section{Implementation Details}
\label{sec:suppl_details}

\subsection{Teacher-Forcing and Self-Forcing}
\label{sec:suppl_forcing}

\begin{figure}[t]
\centering
\includegraphics[width=\linewidth]{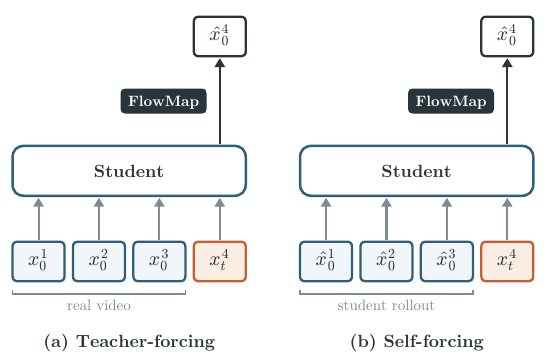}
\caption{\textbf{Teacher-forcing and self-forcing.} The student generates the chunk $\hat{x}^4_0$ from the noisy chunk $x^4_t$ given the history. (a)~In teacher-forcing, the history is clean video $x^{<4}_0$. (b)~In self-forcing, the history is the student's own rollout $\hat{x}^{<4}_0$.}
\label{fig:forcing}
\end{figure}

Figure~\ref{fig:forcing} illustrates how the two stages of AR distillation form the history of each chunk.

\subsection{FlowMap Objectives}
\label{sec:suppl_objectives}

All three objectives use the guided teacher velocity $v^\star$ of Eq.~\eqref{eq:target}. In Stage~2, the clean history $x^{<i}_0$ below is replaced by the student's rollout history $\hat{x}^{<i}_0$.

\paragraph{\piflow}
As an instance of LSD, \piflow \citep{chen2026pi,shaul2026parallel} defines the student flow map through a velocity policy $\pi_\theta(\tau\mid x^i_s,s,x^{<i}_0)$ evaluated once at the source state, $\Phi^\theta_{t\leftarrow s}(x^i_s\mid x^{<i}_0)=x^i_s-\int_t^s\pi_\theta(\tau\mid x^i_s,s,x^{<i}_0)\,d\tau$. Its derivative with respect to $t$ is the policy itself, so the loss needs no Jacobian--vector products:
\begin{equation}
    \mathcal{L}_{\mathrm{LSD}}
    =\mathbb{E}\bigl\|\pi_\theta(t\mid x^i_s,s,x^{<i}_0)-v^\star\bigl(\sg[\hat{x}^i_t],t\mid x^{<i}_0\bigr)\bigr\|^2,
    \label{eq:piflow_loss}
\end{equation}
where $\hat{x}^i_t=\Phi^\theta_{t\leftarrow s}(x^i_s\mid x^{<i}_0)$ and $\sg[\cdot]$ denotes stop-gradient. We use a discrete-velocity policy that predicts one velocity for each of $K=128$ fixed noise-level intervals, so student rollouts are closed-form sums over the intervals between $s$ and $t$ \citep{shaul2026parallel}.

\paragraph{CD}
As an instance of ESD, CD \citep{song2023consistency,zheng2026causal} trains an endpoint map $f_\theta(x^i_s,s\mid x^{<i}_0)=x^i_s-s\,v_\theta(x^i_s,s\mid x^{<i}_0)$ to be consistent along the teacher trajectory. We use its discrete-time form, which keeps $f_\theta$ unchanged across one teacher Euler step $\tilde{x}^i_t=x^i_s-(s-t)\,v^\star(x^i_s,s\mid x^{<i}_0)$:
\begin{equation}
    \mathcal{L}_{\mathrm{ESD}}
    =\mathbb{E}\bigl\|f_\theta(x^i_s,s\mid x^{<i}_0)-\sg\bigl[f_\theta(\tilde{x}^i_t,t\mid x^{<i}_0)\bigr]\bigr\|^2.
    \label{eq:cm_loss}
\end{equation}

\paragraph{MeanFlow}
MeanFlow \citep{geng2026mean} extends ESD to any target noise level $r\le s$ by predicting the average velocity $u_\theta(x^i_s,r,s\mid x^{<i}_0)$ over $[r,s]$, so that $\Phi^\theta_{r\leftarrow s}(x^i_s\mid x^{<i}_0)=x^i_s-(s-r)\,u_\theta$. It is trained toward $v^\star(x^i_s,s\mid x^{<i}_0)-(s-r)\,\sg[\tfrac{d}{ds}u_\theta]$ with adaptive loss weighting. Instead of computing $\tfrac{d}{ds}u_\theta$ with a Jacobian--vector product, we approximate it by a finite difference across one teacher Euler step, as in CD.

\subsection{Uncertainty Head and Normalization}
\label{sec:suppl_uncertainty}

The uncertainty head is a single frame-local Transformer layer on the detached features of block $\ell=10$ of the frozen conditional teacher, and predicts a log-uncertainty $\hat{u}_p$ for each patch $p$ ($2\times2$ latent pixels of one frame). Given a clean chunk $x^i_0$ and clean history $x^{<i}_0$, let $e_p$ be the patch-wise mean squared flow-matching residual $r=v_\phi(x^i_t,t\mid x^{<i}_0)-(x^i_1-x^i_0)$. We train the head with the Gaussian negative log-likelihood $\frac{1}{2}\bigl(e_p\exp(-\hat{u}_p)+\hat{u}_p\bigr)$ \citep{kendall2017uncertainties} for 500 updates with a batch size of 8.

During distillation, we smooth $\hat{u}_p$ with a $3\times3$ mean pool to obtain $\tilde{u}_p$ and normalize it into the skip weight of Eq.~\eqref{eq:soft_skip}, $u_p=\max\bigl(0,\,\bar{u}+\sigma_u(\tilde{u}_p-\mu)/\sigma\bigr)$, with $\bar{u}=\sigma_u=0.5$. Here $\mu$ and $\sigma$ are the mean and standard deviation of $\tilde{u}_p$, precomputed at 64 noise levels for each chunk index and linearly interpolated in between. After clipping at zero, the mean skip weight is $0.61$. We apply uncertainty-weighted soft skips to the negative-prompt branch of CFG (Figure~\ref{fig:motion_guidance}).

\subsection{Drift Model}
\label{sec:suppl_drift_model}

In Stage~2, the drift model $v_\psi$ is initialized from the AR teacher $v_\phi$ and trained online on the student's generated chunks $\hat{x}^i_0$ and rollout histories $\hat{x}^{<i}_0$ ($i>1$) with the flow-matching loss
\begin{equation*}
    \mathcal{L}_\psi=\mathbb{E}\bigl\|v_\psi\bigl((1-t)\hat{x}^i_0+tx^i_1,\,t\mid\hat{x}^{<i}_0\bigr)-(x^i_1-\hat{x}^i_0)\bigr\|^2.
\end{equation*}
After each student update, $v_\psi$ takes one Adam step on the rollout used for that update and $n-1$ steps on fresh rollouts of the updated student, generated without gradients.

\subsection{Training Details}
\label{sec:suppl_training}

\paragraph{\piflow}
Each update samples $M=8$ source states per chunk with a maximum jump of $0.5$. Stage~1 runs 64 teacher-forcing updates with a learning rate of $10^{-5}$. Stage~2 runs 64 self-forcing updates with a learning rate of $2\times10^{-6}$ and Adam $\beta=(0,0.999)$, and the drift model uses a learning rate of $2\times10^{-6}/n$. The batch size is 64 with the Causal-rCM teacher and 32 with the Causal Forcing teacher.

\paragraph{CD and MeanFlow}
Stage~1 runs 2,000 teacher-forcing updates and Stage~2 runs 2,000 self-forcing updates, with $n=1$ drift-model update per student update at a learning rate of $2\times10^{-6}$. CD uses a learning rate of $2\times10^{-6}$ and AdamW $\beta=(0,0.999)$ in both stages, whereas MeanFlow uses a learning rate of $10^{-5}$ with $\beta=(0.9,0.999)$ in Stage~1 and $2\times10^{-6}$ with $\beta=(0,0.999)$ in Stage~2. As in \piflow, the batch size is 64 with the Causal-rCM teacher and 32 with the Causal Forcing teacher. In Stage~2, source states follow the student's own 4-step sampler on its rollout history $\hat{x}^{<i}_0$. Let $s_k$ be the last sampler time with $s_k\ge s$. For CD, whose sampler re-noises an $x_0$ estimate at every step, $x^i_s$ re-noises with fresh noise the $x_0$ estimate that the sampler re-noises at $s_k$. For MeanFlow, whose flow-map sampler is deterministic, $x^i_s=\Phi^\theta_{s\leftarrow s_k}(x^i_{s_k}\mid\hat{x}^{<i}_0)$ is the student's own jump from its sampler state at $s_k$, without re-noising.

\subsection{Evaluation}
\label{sec:suppl_eval}

All models generate one video per prompt and seed, and VBench scores are averaged over seeds 0--4. Diversity is computed on the same videos with the V-JEPA~2 and VideoMAE~V2 encoders, following the protocol of PDD \citep{shaul2026parallel}. For 30\,s videos, all models use the KV-cache management of Deep Forcing \citep{yi2025deep} (Deep Sink and participative compression) with the same settings, and VBench-Long scores are averaged over the 16 clips of each video.

\section{Additional Results}
\label{sec:suppl_results}

\subsection{Motion-Aware Guidance in Stage~1}
\label{sec:suppl_stage1}

Table~\ref{tab:abl_stage1} compares \methodtwo with CFG alone and with uniform skip-layer guidance after Stage~1. For \piflow, uniform skip-layer guidance degrades quality at $u=0.5$ and $u=1$, and nearly matches CFG alone when the skip weight is set to the mean skip weight of \methodtwo ($u=0.61$: 83.69 vs.\ 83.66), whereas \methodtwo reaches 84.01. Motion-aware guidance improves VBench for both \piflow and CD, showing that the gain comes from where the skip is applied rather than from its average strength.

\begin{table}[t]
\centering
\caption{\textbf{\Methodtwo in Stage~1 (Causal Forcing teacher).} Students evaluated after Stage~1. Fixed variants apply skip-layer guidance with a uniform skip weight $u$ across all patches. Here $u=0.5$ matches the base skip weight of \methodtwo, and $u=0.61$ matches its mean skip weight during training. Removing skip-layer guidance leaves CFG alone.}
\label{tab:abl_stage1}
\small
\begin{tabular}{llc}
\toprule
Method & Teacher guidance & VBench $\uparrow$ \\
\midrule
\piflow & \cellcolor{gray!15}Motion-aware (ours) & \cellcolor{gray!15}\textbf{84.01} \\
 & \quad fixed $u=0.5$ & 83.54 \\
 & \quad fixed $u=0.61$ & 83.69 \\
 & \quad fixed $u=1$ & 83.43 \\
 & \quad $-$ skip-layer guidance & 83.66 \\
\midrule
CD & \cellcolor{gray!15}Motion-aware (ours) & \cellcolor{gray!15}\textbf{82.88} \\
 & \quad $-$ skip-layer guidance & 82.64 \\
\bottomrule
\end{tabular}
\end{table}

\subsection{\Methodone: Sample Diversity}
\label{sec:suppl_drift}

Table~\ref{tab:suppl_drift} repeats Table~\ref{tab:abl_gamma} with the Causal-rCM teacher. While $n=1$ slightly lowers diversity relative to disabling \methodone ($\gamma=1$), our default setting $n=5$ achieves the highest diversity under both V-JEPA~2 and VideoMAE~V2 while keeping 5\,s VBench close to $\gamma=1$, confirming that \methodone with an accurately tracked drift model ($n=5$) does not hurt sample diversity.

\begin{table}[t]
\centering
\caption{\textbf{Drift-model update frequency $n$} on \piflow with the Causal-rCM teacher at $\gamma=5$ ($\gamma=1$: no \methodone).}
\label{tab:suppl_drift}
\small
\setlength{\tabcolsep}{4pt}
\begin{tabular}{ccccc}
\toprule
$n$ & $\gamma$ & VBench $\uparrow$ & V-JEPA 2 $\uparrow$ & VideoMAE V2 $\uparrow$ \\
\midrule
\rowcolor{gray!15} 5 & 5 (ours) & 84.87 & \textbf{0.0735} & \textbf{0.0143} \\
1 & 5 & 84.71 & 0.0703 & 0.0130 \\
-- & 1 & \textbf{84.90} & 0.0708 & 0.0136 \\
\bottomrule
\end{tabular}
\end{table}

\subsection{\Methodone: Long-Video Generation}
\label{sec:suppl_stage2}

Table~\ref{tab:suppl_stage2} repeats the \piflow rows of Table~\ref{tab:abl_stage2} with the Causal-rCM teacher. \Methodone again reduces the 30\,s quality drop, on both $\Delta$IQ and $\Delta$AQ, while the 30\,s dynamic degree stays comparable.

\begin{table}[t]
\centering
\caption{\textbf{\Methodone in Stage~2 (Causal-rCM teacher).} All variants use \methodtwo in both stages; removing \methodone sets $\gamma=1$. All metrics are measured on 30\,s rollouts.}
\label{tab:suppl_stage2}
\footnotesize
\setlength{\tabcolsep}{2pt}
\begin{tabular}{llccc}
\toprule
Method & Teacher guidance & DD $\uparrow$ & $\Delta$IQ $\uparrow$ & $\Delta$AQ $\uparrow$ \\
\midrule
\piflow & \cellcolor{gray!15}Anti-drifting (ours) & \cellcolor{gray!15}41.1 & \cellcolor{gray!15}\textbf{$-$1.64} & \cellcolor{gray!15}\textbf{$-$3.83} \\
 & \quad $-$ \methodone & \textbf{42.0} & $-$3.88 & $-$6.41 \\
\bottomrule
\end{tabular}
\end{table}

\subsection{All VBench-Long Dimensions}
\label{sec:suppl_long}

Table~\ref{tab:suppl_long} lists all VBench-Long dimensions of the 30\,s videos in Table~\ref{tab:long}.

\begin{table*}[t]
\centering
\caption{\textbf{All VBench-Long dimensions on 30\,s videos.} Results for the chunk-wise 4-NFE students of Table~\ref{tab:long}, averaged over five seeds (KV-cache pattern follows \citet{yi2025deep}). Overall denotes overall text-video consistency, and $\Delta$IQ and $\Delta$AQ are the changes in imaging and aesthetic quality from the first three to the last three of 16 clips of about 2\,s. Colors follow Table~\ref{tab:main}.}
\label{tab:suppl_long}
\footnotesize
\setlength{\tabcolsep}{3.2pt}
\begin{tabular}{lccccccccc}
\toprule
Method & Subject $\uparrow$ & Background $\uparrow$ & Smoothness $\uparrow$ & Dynamic $\uparrow$ & Aesthetic $\uparrow$ & Imaging $\uparrow$ & Overall $\uparrow$ & $\Delta$IQ $\uparrow$ & $\Delta$AQ $\uparrow$ \\
\midrule
Self Forcing$^{\dagger}$ & 96.34 & 95.84 & 98.11 & 60.8 & 61.18 & 69.30 & 20.59 & $-$0.76 & $-$2.00 \\
\midrule
\multicolumn{10}{l}{\textit{Causal-rCM teacher}} \\
Causal-rCM$^{\dagger}$ & 96.18 & 95.78 & 98.35 & \vbbest{65.6} & \vbthird{61.41} & 65.92 & \vbthird{20.52} & $-$3.28 & \vbthird{$-$4.06} \\
Causal-rCM, TF-sCM$^{\dagger}$ & 95.94 & 96.06 & 97.85 & \vbsecond{48.6} & 58.80 & 65.29 & \vbsecond{20.78} & $-$3.10 & $-$5.08 \\
Ours (\piflow) & \vbbest{98.07} & \vbbest{97.24} & \vbbest{98.87} & 41.1 & \vbbest{62.13} & \vbsecond{67.41} & 20.18 & \vbbest{$-$1.64} & \vbbest{$-$3.83} \\
Ours (MeanFlow) & \vbthird{97.69} & \vbsecond{96.61} & \vbthird{98.80} & \vbthird{41.7} & \vbsecond{61.81} & \vbthird{67.19} & \vbbest{21.00} & \vbsecond{$-$1.95} & $-$4.12 \\
Ours (CD) & \vbsecond{97.73} & \vbthird{96.49} & \vbsecond{98.86} & 41.3 & 60.27 & \vbbest{68.25} & 20.41 & \vbthird{$-$1.97} & \vbsecond{$-$4.01} \\
\midrule
\multicolumn{10}{l}{\textit{Causal Forcing teacher}} \\
Causal Forcing$^{\dagger}$ & \vbthird{95.01} & \vbsecond{94.73} & 97.00 & \vbsecond{81.7} & \vbbest{58.71} & \vbbest{68.92} & \vbsecond{19.42} & \vbbest{$-$1.44} & \vbbest{$-$5.59} \\
Ours (\piflow) & 94.78 & \vbthird{94.50} & \vbthird{97.90} & \vbbest{82.1} & \vbthird{57.92} & \vbsecond{64.62} & 18.82 & \vbsecond{$-$4.94} & \vbsecond{$-$6.80} \\
Ours (MeanFlow) & \vbsecond{95.91} & \vbbest{95.15} & \vbsecond{98.38} & \vbthird{59.8} & \vbsecond{58.37} & \vbthird{63.12} & \vbbest{19.69} & \vbthird{$-$6.50} & \vbthird{$-$8.47} \\ 
Ours (CD) & \vbbest{96.03} & 94.42 & \vbbest{98.79} & 49.4 & 54.89 & 62.04 & \vbthird{19.00} & $-$9.01 & $-$11.56 \\
\bottomrule
\end{tabular}
\end{table*}

\subsection{More Qualitative Results}
\label{sec:suppl_qual}

\begin{figure*}[p]
\centering
\includegraphics[width=\linewidth]{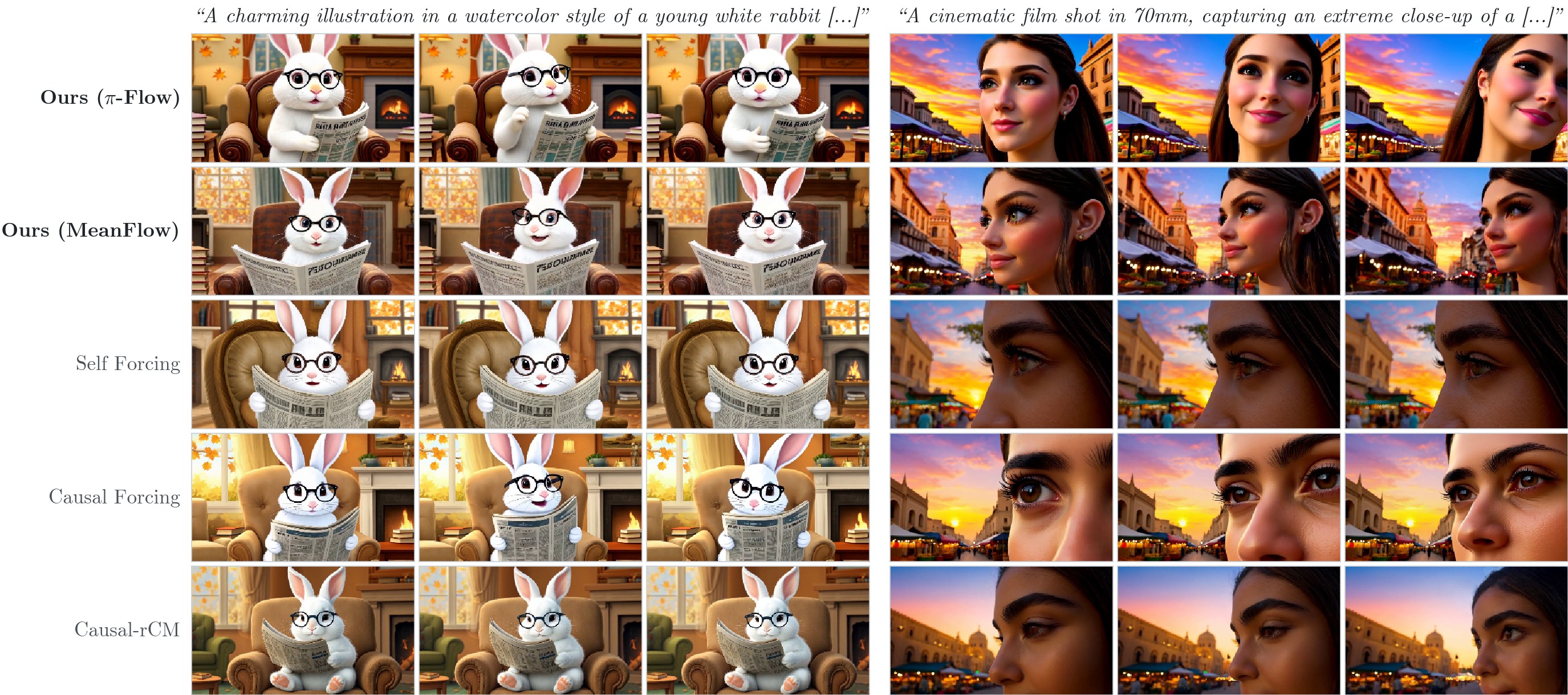}\\[6pt]
\includegraphics[width=\linewidth]{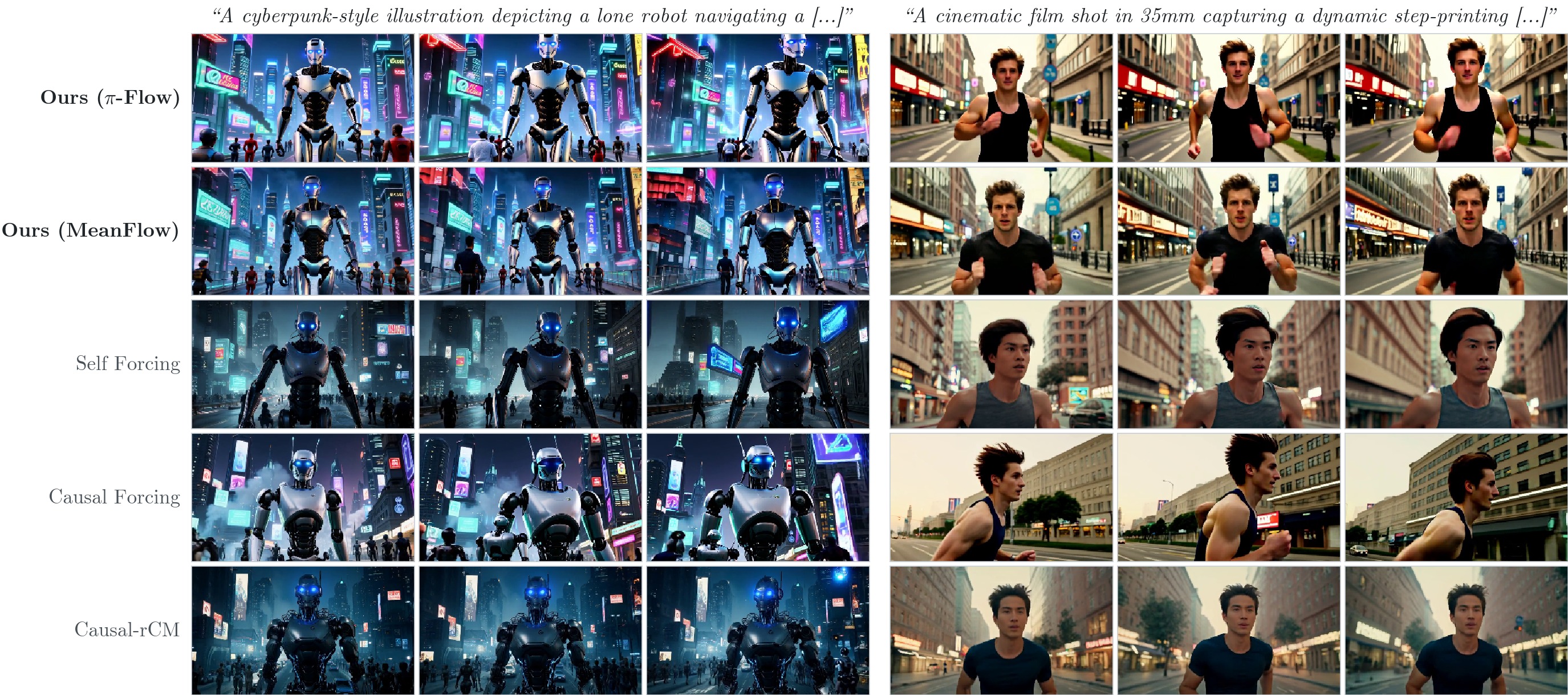}
\caption{\textbf{More qualitative comparisons on Wan2.1-1.3B (1/3).} Frames are shown at $t\in\{0.5, 2.5, 4.5\}$\,s. Prompts are shortened to fit one line. Ours are distilled from the Causal Forcing teacher.}
\label{fig:gallery_hq_a}
\end{figure*}

\begin{figure*}[p]
\centering
\includegraphics[width=\linewidth]{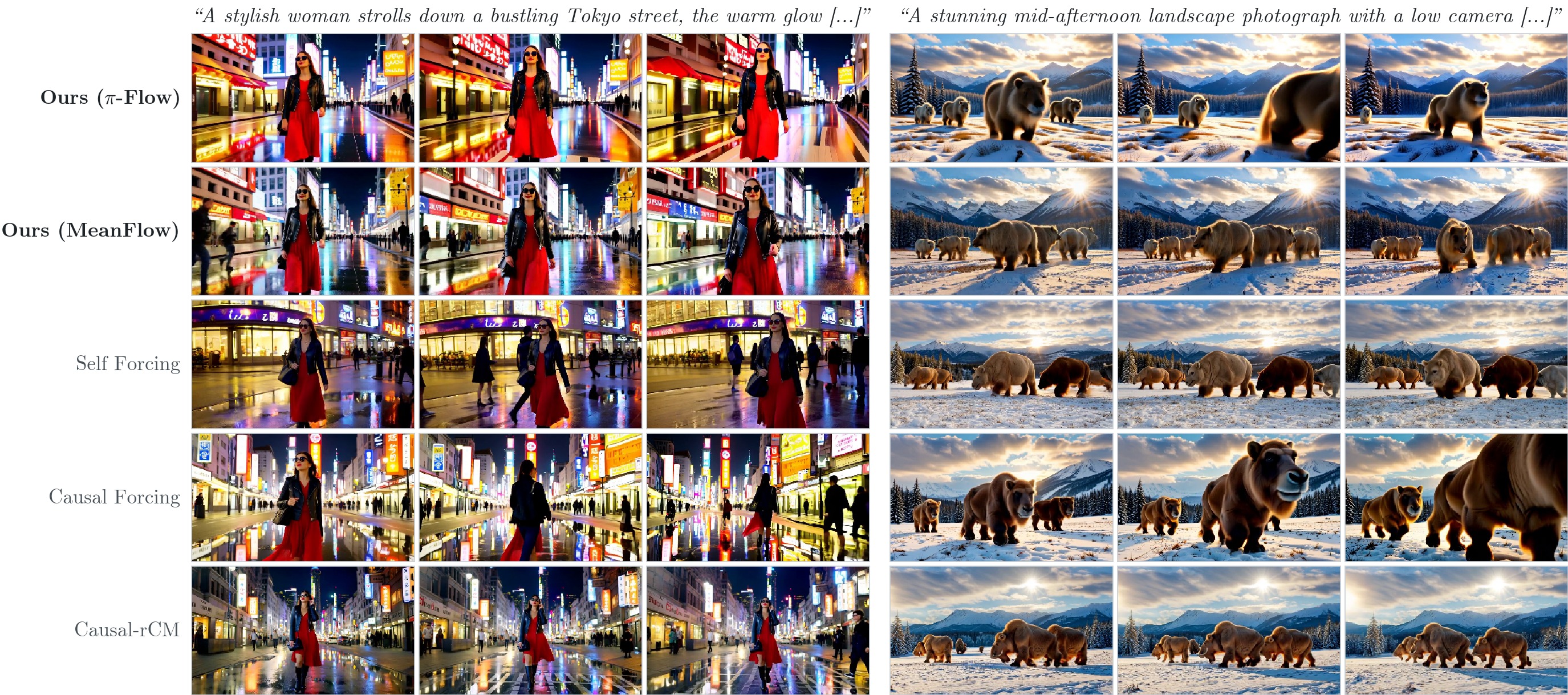}\\[6pt]
\includegraphics[width=\linewidth]{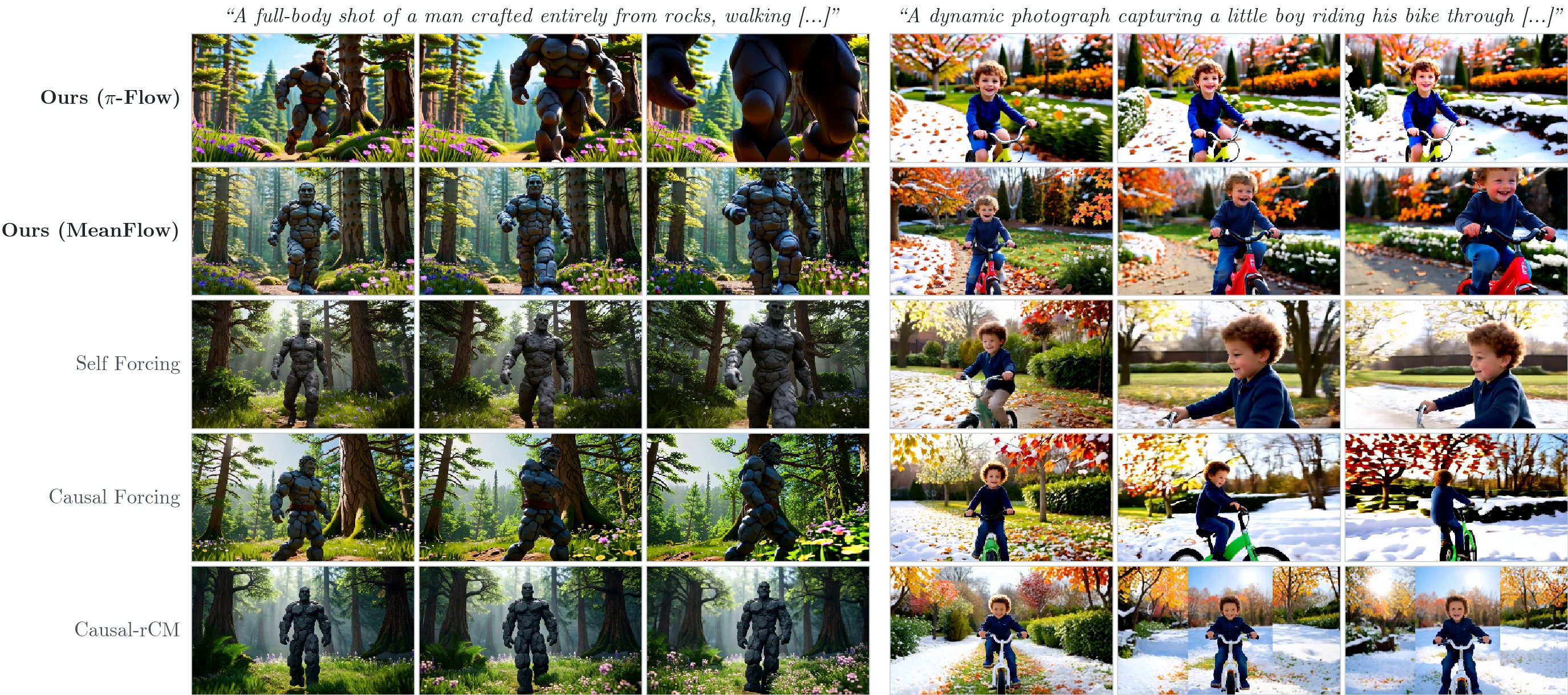}
\caption{\textbf{More qualitative comparisons on Wan2.1-1.3B (2/3).}}
\label{fig:gallery_hq_b}
\end{figure*}

\begin{figure*}[p]
\centering
\includegraphics[width=\linewidth]{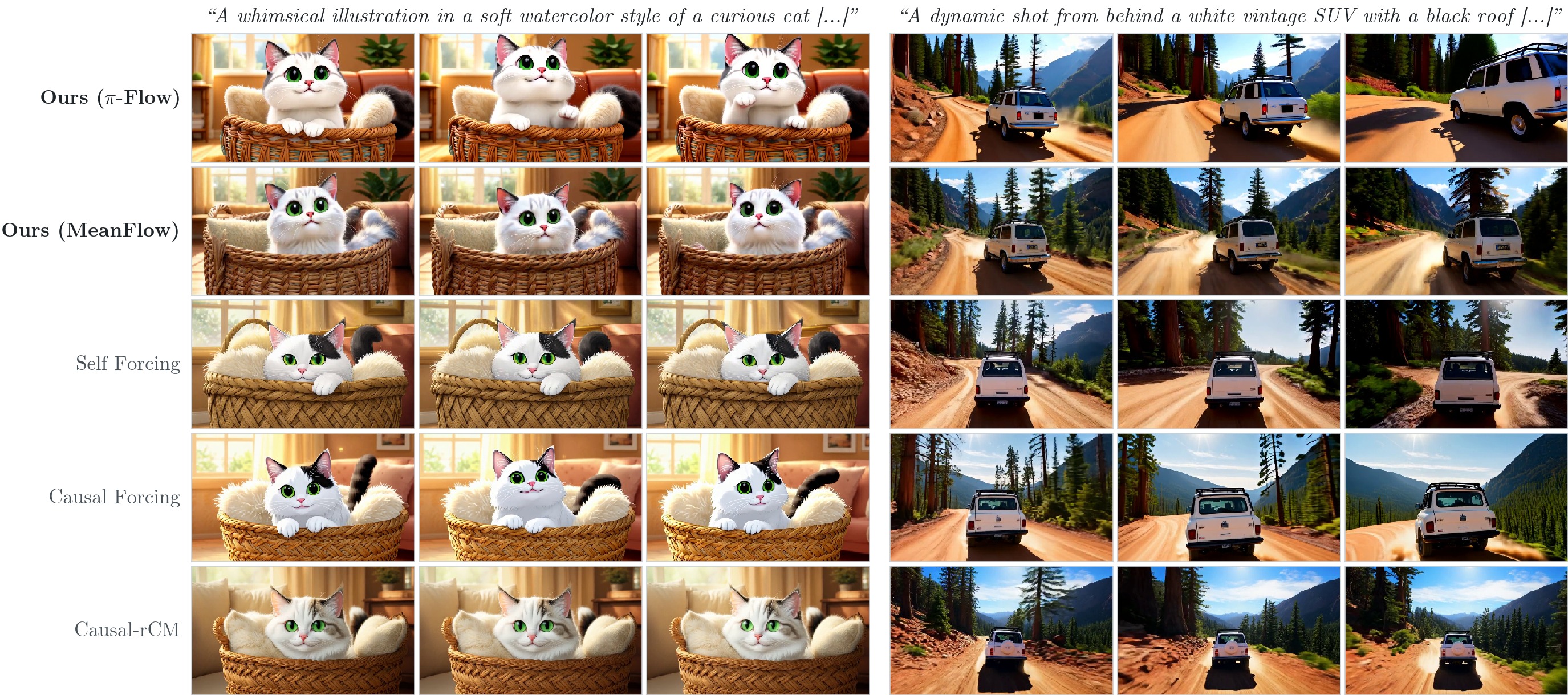}
\caption{\textbf{More qualitative comparisons on Wan2.1-1.3B (3/3).}}
\label{fig:gallery_hq_c}
\end{figure*}

\begin{figure*}[p]
\centering
\includegraphics[width=\linewidth]{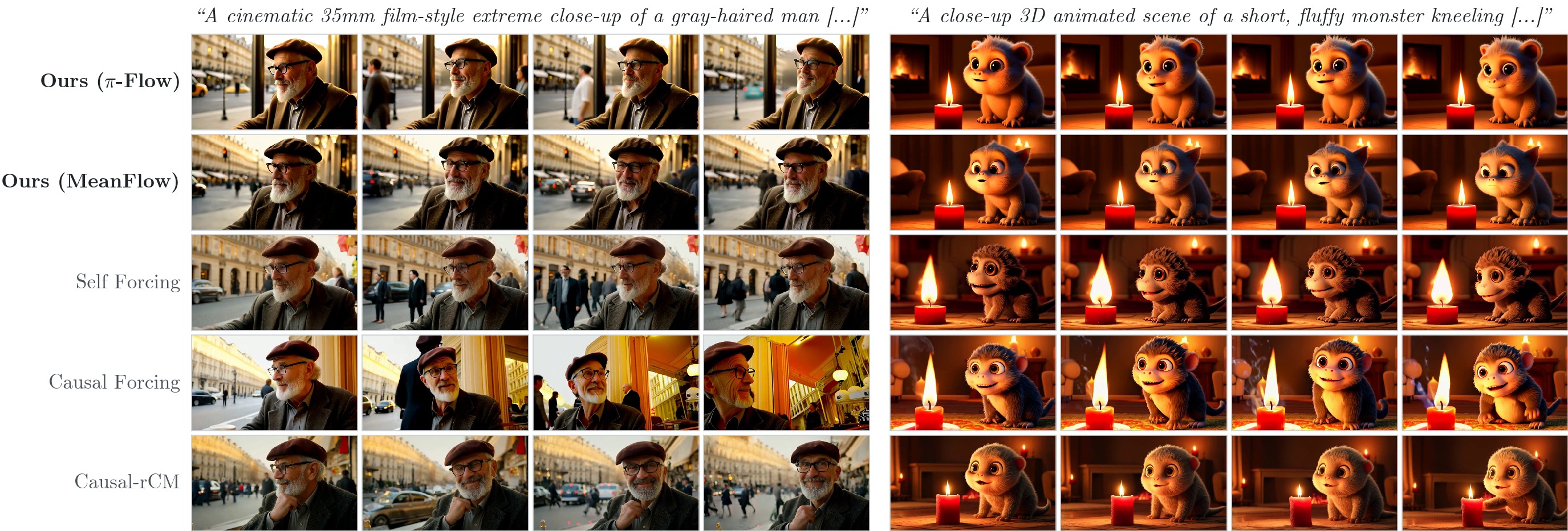}\\[6pt]
\includegraphics[width=\linewidth]{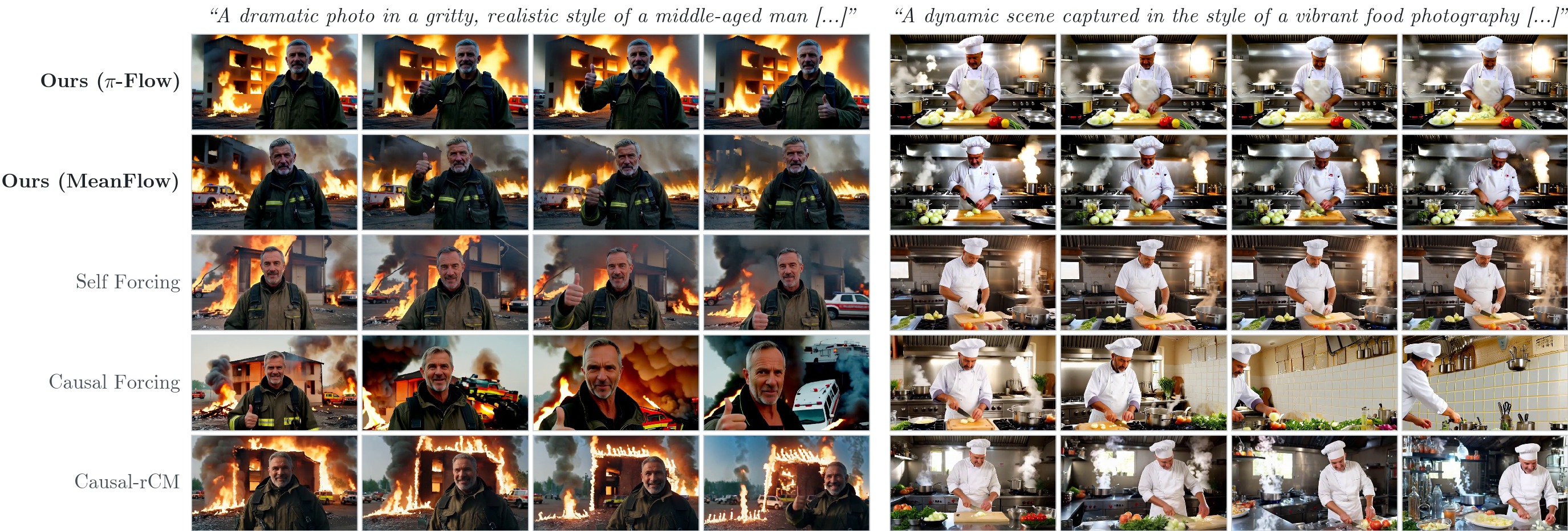}\\[6pt]
\includegraphics[width=\linewidth]{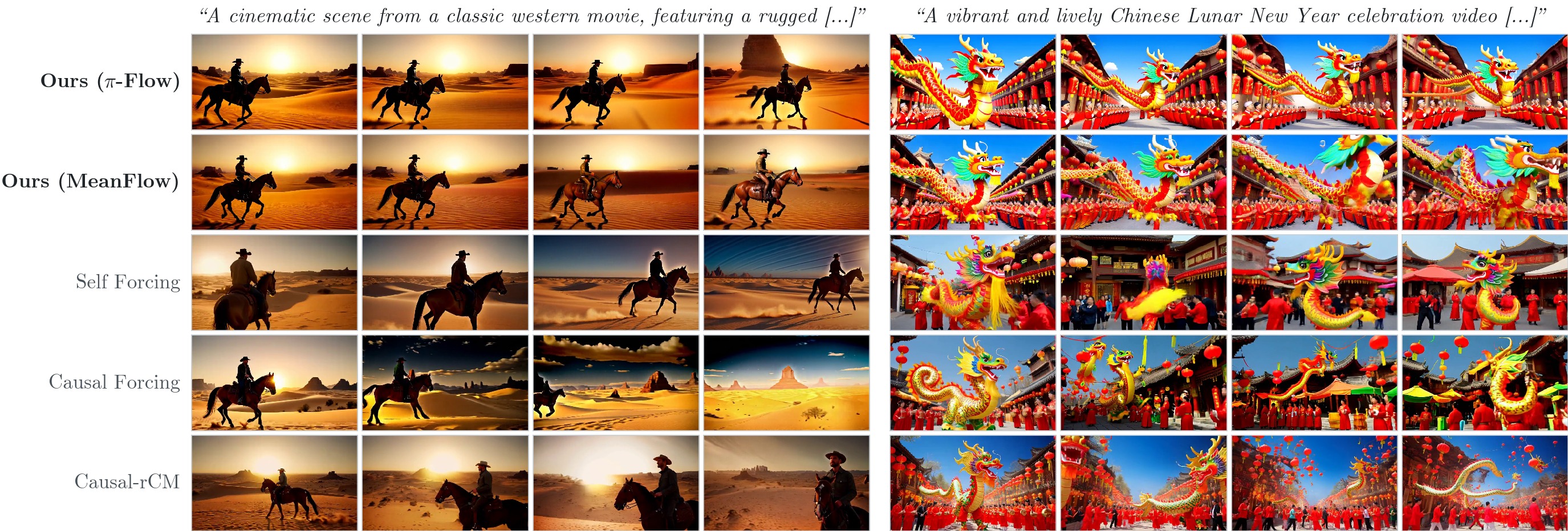}
\caption{\textbf{More 30\,s comparisons on Wan2.1-1.3B (1/2).} Frames are shown at $t\in\{0, 10, 20, 30\}$\,s. Prompts are shortened to fit one line. Ours are distilled from the Causal-rCM teacher, and all methods use the KV-cache management of \citet{yi2025deep}.}
\label{fig:gallery_long_a}
\end{figure*}

\begin{figure*}[p]
\centering
\includegraphics[width=\linewidth]{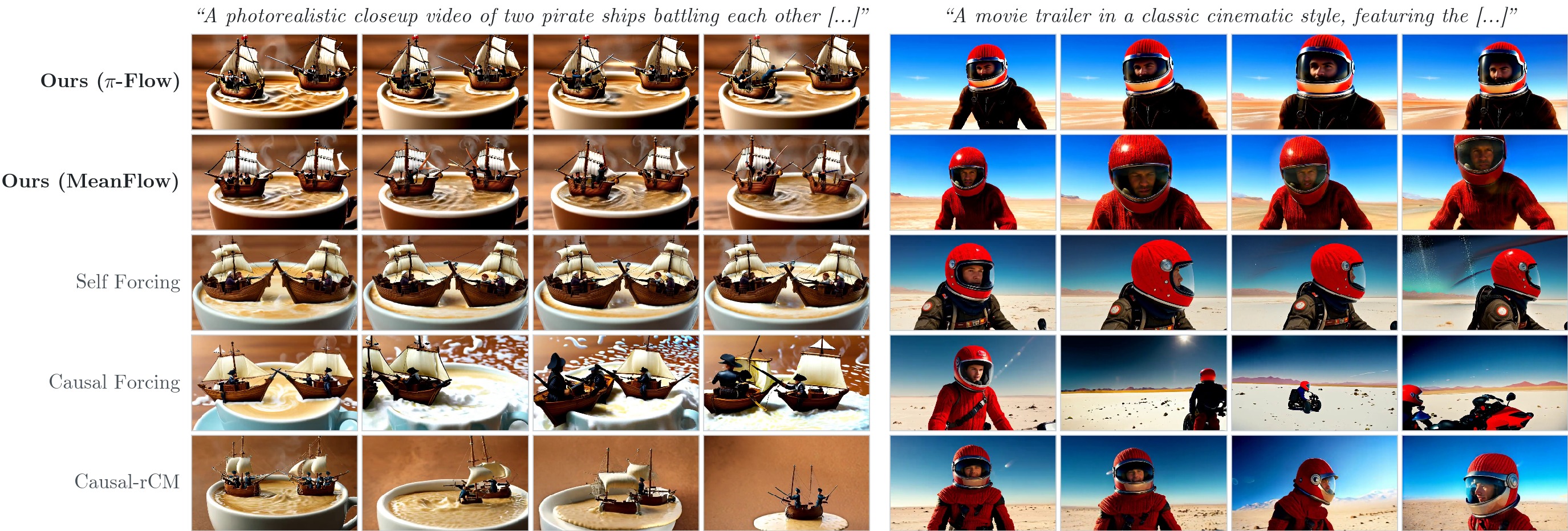}\\[6pt]
\includegraphics[width=\linewidth]{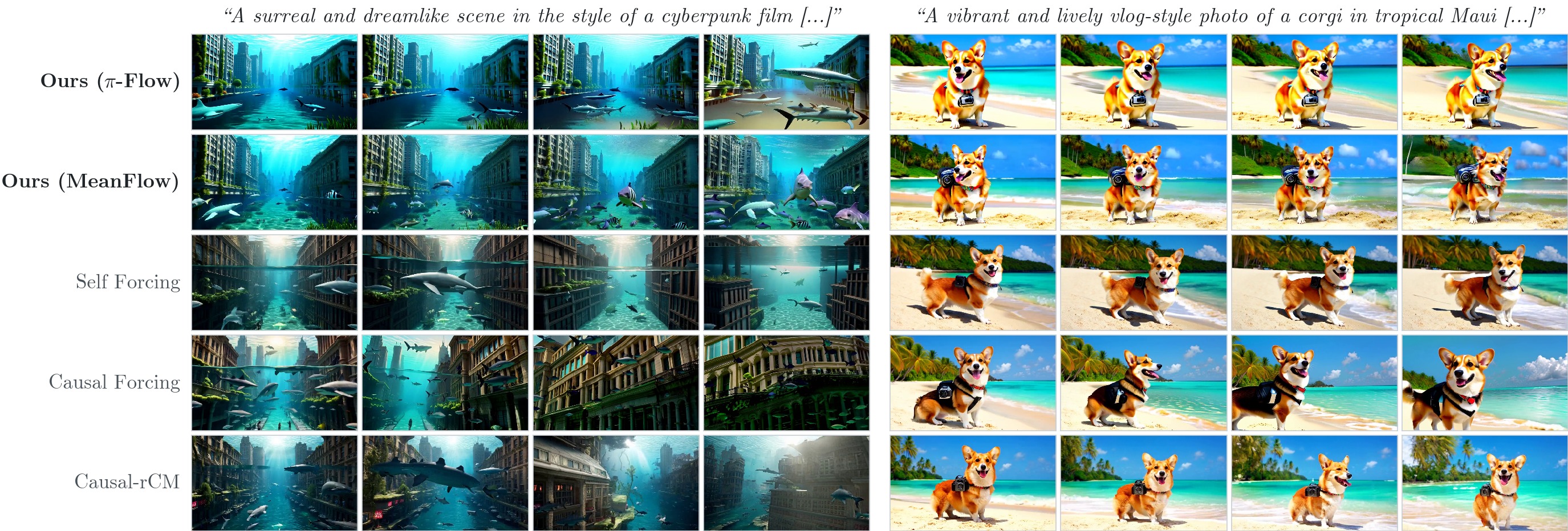}
\caption{\textbf{More 30\,s comparisons on Wan2.1-1.3B (2/2).}}
\label{fig:gallery_long_b}
\end{figure*}

\begin{figure*}[p]
\centering
\includegraphics[width=\linewidth]{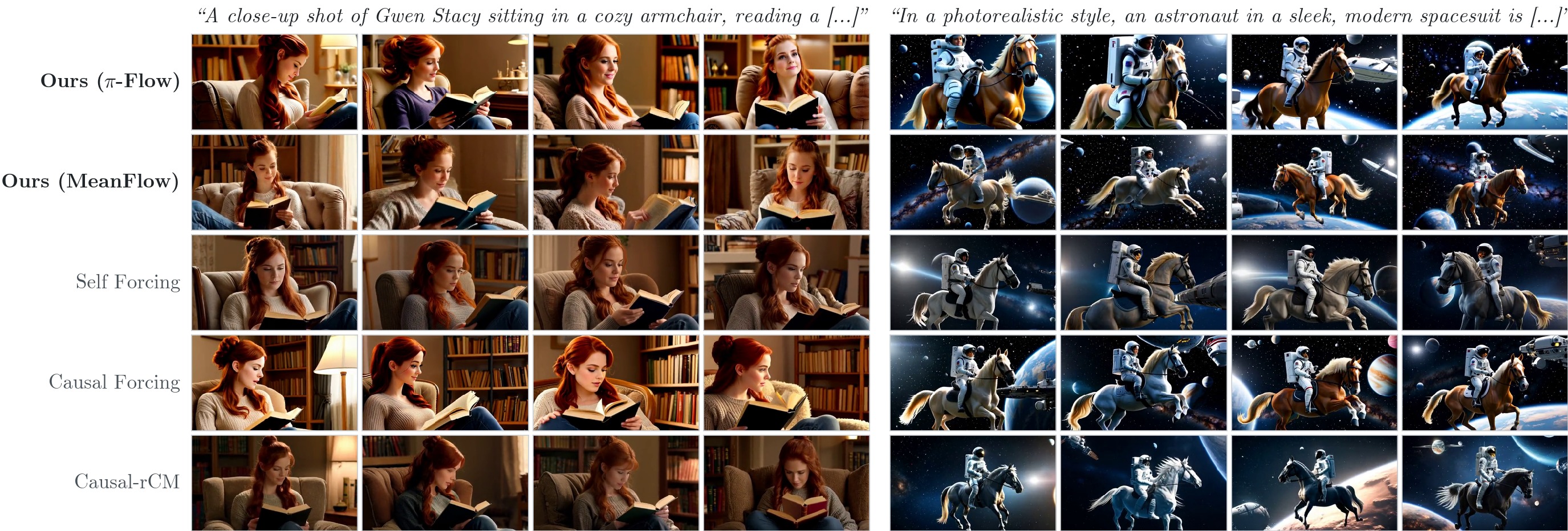}\\[6pt]
\includegraphics[width=\linewidth]{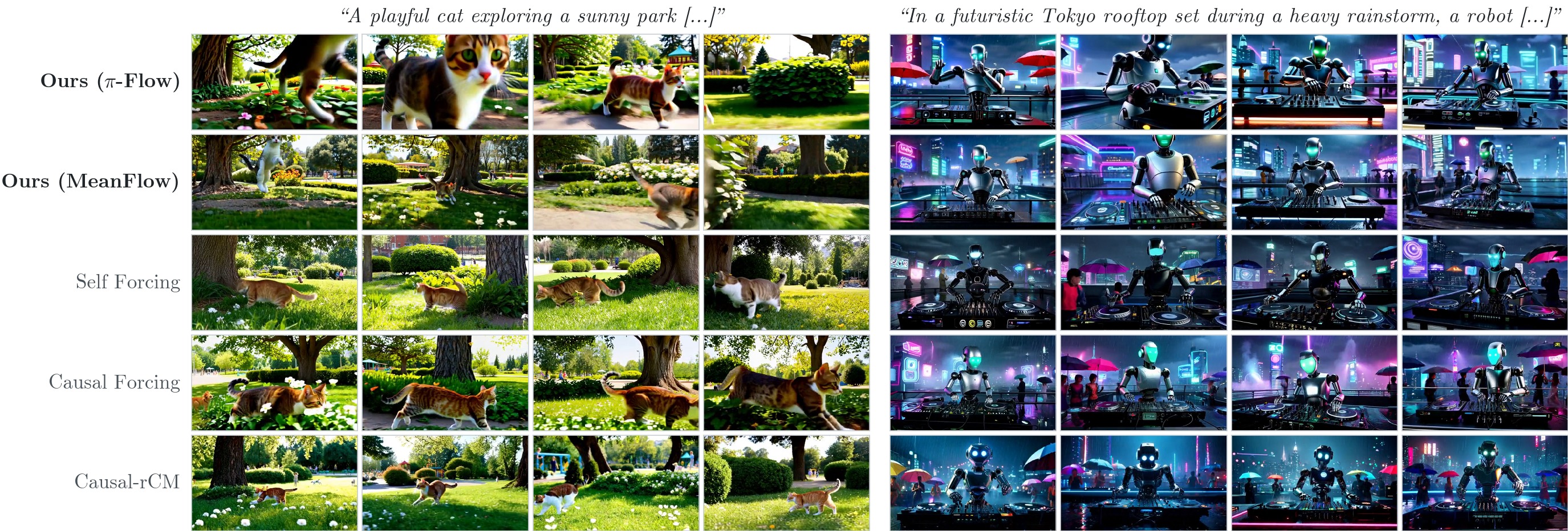}\\[6pt]
\includegraphics[width=\linewidth]{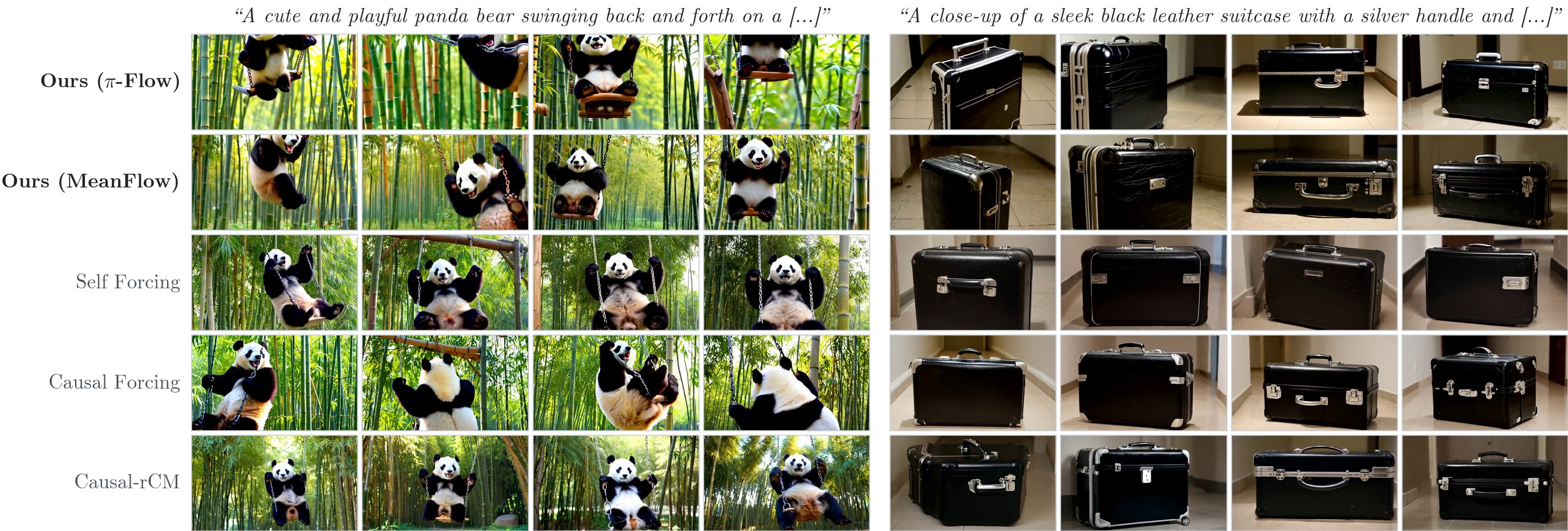}
\caption{\textbf{More diversity comparisons on Wan2.1-1.3B (1/2).} Each frame is taken at 2.5\,s from a video with a different seed. Ours are distilled from the Causal Forcing teacher.}
\label{fig:gallery_div_a}
\end{figure*}

\begin{figure*}[p]
\centering
\includegraphics[width=\linewidth]{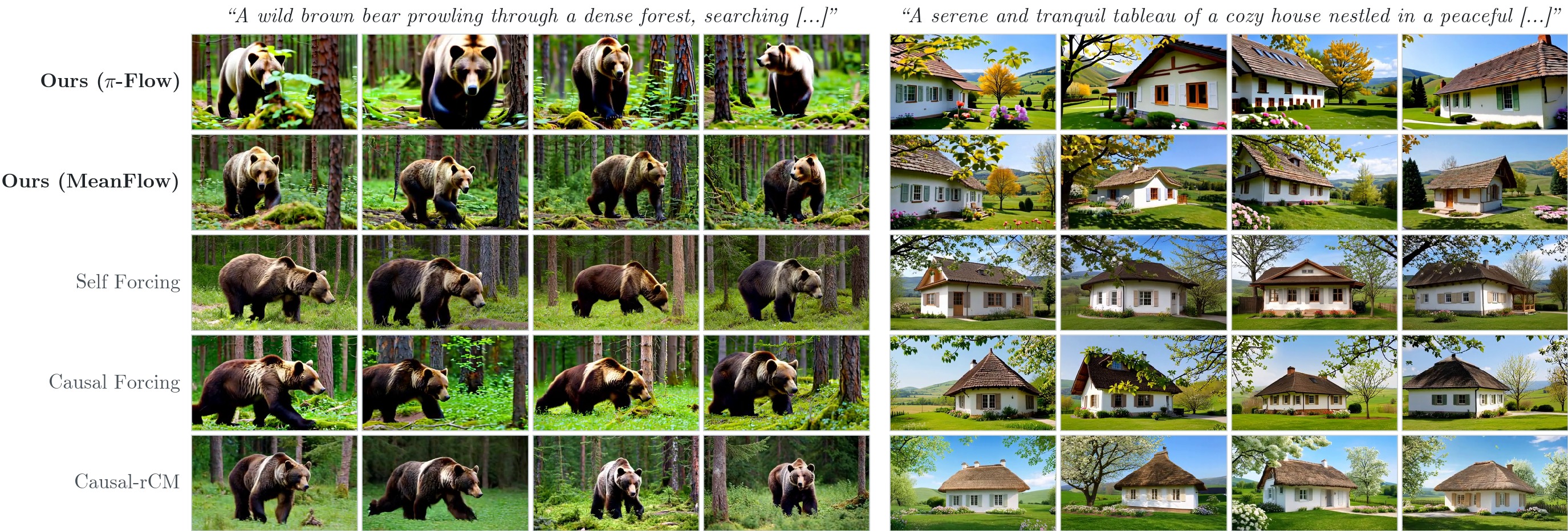}\\[6pt]
\includegraphics[width=\linewidth]{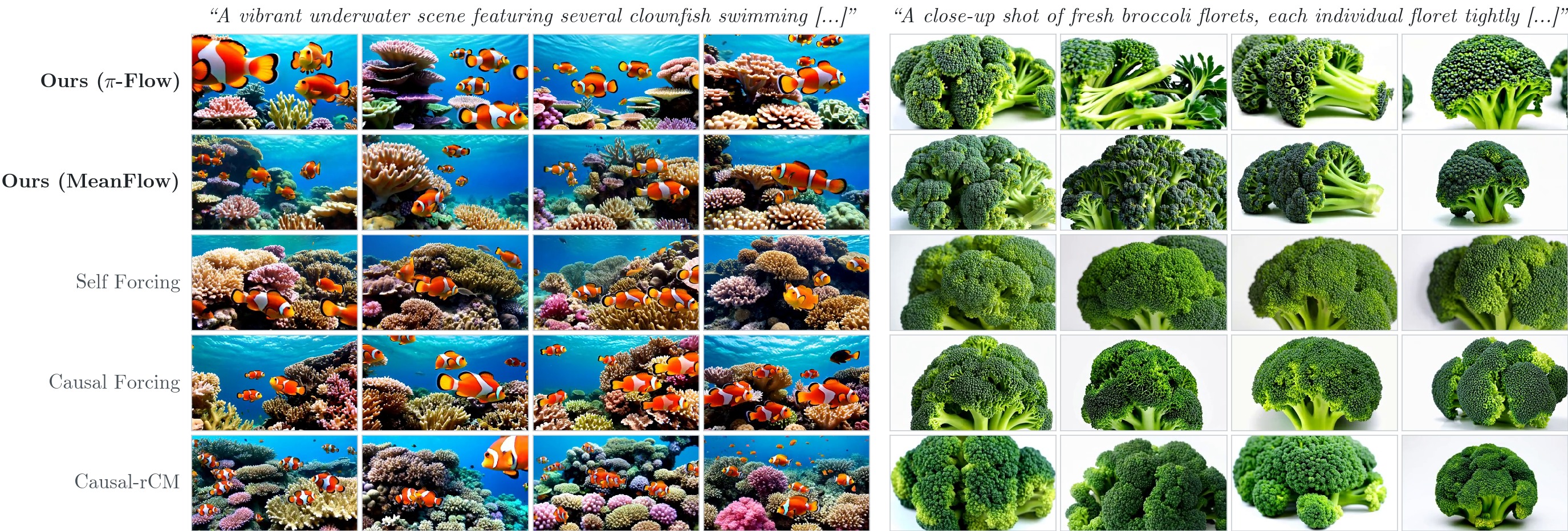}
\caption{\textbf{More diversity comparisons on Wan2.1-1.3B (2/2).}}
\label{fig:gallery_div_b}
\end{figure*}

Figures~\ref{fig:gallery_hq_a}--\ref{fig:gallery_div_b} show more examples of 5\,s generation, 30\,s generation, and sample diversity, in the same layout as Figures~\ref{fig:qualitative}, \ref{fig:longvideo}, and~\ref{fig:diversity}. Within each panel, all methods use the same prompt and seed.

\paragraph{Example selection}
Prompts for 5\,s and 30\,s generation come from MovieGen Bench, and prompts for diversity come from the extended VBench prompts. We chose the examples in all qualitative figures by visual inspection of our students' outputs, and show the baselines for the same prompt and seed.

\end{document}